%% file: main.tex
\pdfoutput=1

\documentclass[11pt]{article}

\usepackage[]{acl}

\usepackage{times}
\usepackage{latexsym}
\usepackage{hyperref}
\usepackage{booktabs}
\usepackage{amsfonts}
\usepackage{xcolor,colortbl}
\definecolor{amethyst}{rgb}{0.6, 0.4, 0.8}
\definecolor{apricot}{rgb}{0.98, 0.81, 0.69}
\definecolor{atomictangerine}{rgb}{1.0, 0.6, 0.4}
\definecolor{orange_assign}{rgb}{0.9686274509803922, 0.8509803921568627, 0.7686274509803922}
\usepackage{array}
\newcolumntype{P}[1]{>{\centering\arraybackslash}p{#1}}
\usepackage{longtable}
\usepackage{makecell}
\usepackage{multirow}
\usepackage[T1]{fontenc}
\usepackage{enumitem}

\usepackage[utf8]{inputenc}
\usepackage[inkscapelatex=false]{svg}
\usepackage{tablefootnote}

\usepackage{microtype}
\usepackage{amsmath}
\usepackage{listings}
\usepackage{tabularx}

\newcommand{\frameworkname}{TopicGPT}

\title{\frameworkname: A Prompt-based Topic Modeling Framework}
\author{
    Chau Minh Pham$^{1}$ \quad
    Alexander Hoyle$^{2}$ \quad
    Simeng Sun$^{1}$ \quad
    Philip Resnik$^{2}$ \quad
    Mohit Iyyer$^{1}$ \\[0.5em] 
    $^{1}$University of Massachusetts Amherst \quad $^{2}$University of Maryland\\
    \texttt{\{ctpham, simengsun, miyyer\}@umass.edu}, \\
    \texttt{\{hoyle, resnik\}@umd.edu}
}
\begin{document}

\maketitle
\def\thefootnote{\arabic{footnote}}
\input{sections/0-Abstract}
\input{sections/1-Introduction}
\input{sections/2-Related-Work}
\input{sections/3-Framework}

\input{sections/4-Experiments}

\input{sections/5-Results}
\input{sections/6-Discussion}
\input{sections/7-Conclusion}
\input{sections/Limitation}

\bibliography{anthology}
\bibliographystyle{acl_natbib}
\newpage
\input{sections/Appendix}

\end{document}

%% file: sections/0-Abstract.tex
\begin{abstract}
    Topic modeling is a well-established technique for exploring text corpora. Conventional topic models (e.g., LDA) represent topics as bags of words that often require ``reading the tea leaves'' to interpret; additionally, they offer users minimal control over the formatting and specificity of resulting topics. To tackle these issues, we introduce \frameworkname, a prompt-based framework that uses large language models (LLMs) to uncover latent topics in a text collection. \frameworkname~produces topics that align better with human categorizations compared to competing methods: it achieves a harmonic mean purity of 0.74 against human-annotated Wikipedia topics compared to 0.64 for the strongest baseline. Its topics are also interpretable, dispensing with ambiguous bags of words in favor of topics with natural language labels and associated free-form descriptions. Moreover, the framework is highly adaptable, allowing users to specify constraints and modify topics without the need for model retraining. By streamlining access to high-quality and interpretable topics, \frameworkname~represents a compelling, human-centered approach to topic modeling.\footnote{Code at \href{https://github.com/chtmp223/topicGPT}{https://github.com/chtmp223/topicGPT}}
\end{abstract}

%% file: sections/1-Introduction.tex
\section{Introduction}
Topic modeling is a commonly used technique for discovering latent thematic structures in extensive collections of text documents. Traditional topic models such as latent Dirichlet allocation \cite[][LDA]{blei2003latent} represent documents as mixtures of topics, where each topic is a distribution over words.
Topics are often represented with their most probable words, but this representation can contain incoherent or unrelated words that make topics difficult for users to interpret \cite{chang2009reading, newman2010automatic}. Although some models enable users to interactively guide topics based on needs and domain knowledge \cite{hu2014interactive, nikolenko2017topic}, their usability is constrained by the bag-of-words topic format.

\begin{figure*}[ht!]
\centerline{\includegraphics[width=0.9\linewidth]{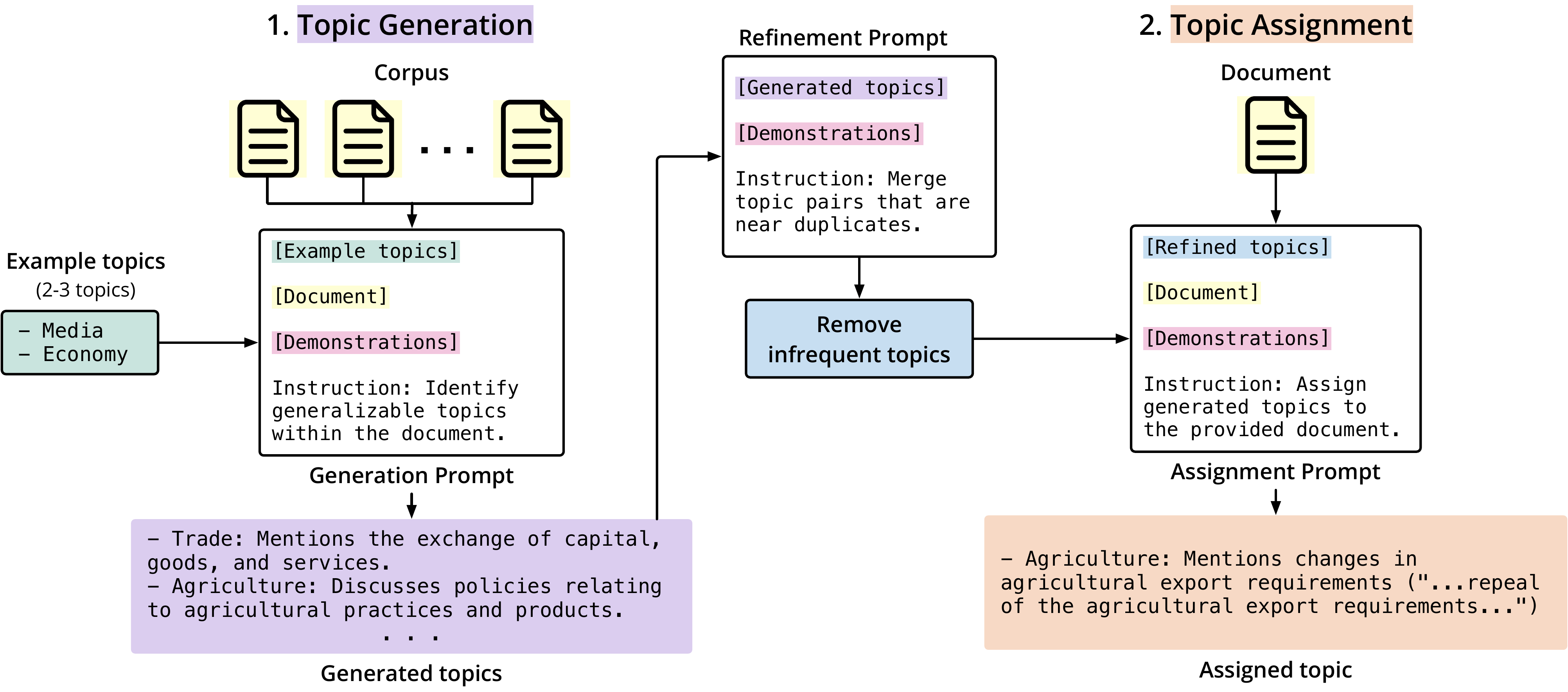}}
\caption{Overview of \frameworkname. \textbf{\textcolor{amethyst}{1) Topic Generation}}: Given a corpus and some manually-curated example topics, \frameworkname~identifies additional topics in each corpus document. The framework then refines the list by merging repeated topics and removing infrequent topics. \textbf{\textcolor{atomictangerine}{2) Topic Assignment}}: Given the generated topics, \frameworkname~assigns the most relevant topic to each document and provides a quote that supports this assignment.}
\label{fig:pipeline}
\end{figure*}

To address these limitations, we introduce \frameworkname~(Figure \ref{fig:pipeline}), a human-centric approach to topic modeling that relies on prompting large language models to perform in-context topic generation (\S \ref{subsec:proposer}) and assignment (\S \ref{subsec:assigner}). First, we iteratively prompt an LLM to generate new topics given a sample of documents from an input dataset and a list of previously generated topics. The resulting set of topics can then be refined to eliminate redundant and infrequent topics. Finally, given a new document, an LLM assigns it to one or more of the generated topics, also providing a quotation from the document to support its assignment. These quotations make the method easily verifiable, addressing some validity concerns plaguing traditional topic models.

\paragraph{\frameworkname\ produces higher-quality topics than competing approaches.} \frameworkname's topics and assignments align significantly more closely with human-annotated ground truth topics than those from LDA, SeededLDA \cite{jagarlamudi_incorporating_2012}, and BERTopic \cite{grootendorst2022bertopic} on two datasets: Wikipedia articles \cite{MerityWiki} and Congressional bills~\cite[as processed by][]{hoyle-etal-2022-neural}. We measure topical alignment using three external clustering metrics (harmonic mean purity, normalized mutual information, and adjusted Rand index) and find that \frameworkname~improves substantially over baselines (e.g., absolute purity improves by 10 points over LDA on Wikipedia); furthermore, its topics are more semantically aligned with human-labeled topics (30.3\% of \frameworkname's topics are misaligned compared to 62.4\% for LDA on Wikipedia). Further analyses demonstrate the robustness of \frameworkname's topic quality across various prompt and data settings.

\paragraph{\frameworkname\ produces more interpretable topics.}
Topics generated by \frameworkname~include natural language labels and descriptions that make them immediately interpretable without needing a separate labeling step. Furthermore, the framework provides informative document-topic associations along with contextual quoted evidence. By creating intuitive topic structures and understandable document-topic assignments, \frameworkname~aims to make the overall process interpretable.

\paragraph{\frameworkname\ is customizable to fit user needs:}
Topic models should suit their application settings \cite{doogan-buntine-2021-topic}. In addition to interpretability, \frameworkname~offers users the ability to guide the generated topics to their specific needs. 
To steer the scope and formatting of the generated topics, users initially provide a small number of example topics, which need not be data-specific or representative of all possible topics in the corpus. After reviewing the initial results, users can manually edit or remove any topics to curate a coherent and goal-oriented list. 

\paragraph{Open-source LLMs are competent topic assigners but bad topic generators.} 
Most of our experiments implement topic generation with GPT-4 \cite{openai2023gpt4} and topic assignment with GPT-3.5-turbo, which costs around \$100 per dataset (Table \ref{tab:cost}). To remove the dependence on expensive APIs, we also experiment with the open-source Mistral-7B-Instruct model \cite{jiang2023mistral}, which assigns topics well but cannot competently follow instructions for generating topics. Thus, improving topic generation in open-source LLMs is an important direction for future work.

%% file: sections/2-Related-Work.tex
\section{Related Work}
We designed \frameworkname\ for the use case of automated content analysis, which is one of the primary applications of topic models \cite{hoyle-etal-2022-neural}. 
In this methodology, practitioners first engage in open coding of individual documents to establish initial labels. These initial codes are then carefully examined, reconfigured, and organized into a cohesive coding system \cite{hsieh2005three, kyngas2020inductive, vears2022inductive}.

\paragraph{Topic modeling for content analysis:} 
Traditional approaches to topic modeling, such as Latent Dirichlet Allocation \cite[][LDA]{blei2003latent}, are parameterized by topic-word and document-topic distributions that can reveal latent thematic structures in a corpus. However, these representations are not straightforward to interpret \cite{Mei2007AutomaticLO, chang2009reading, ramage_partially_2011}, and subjective manual effort may result in issues of reliability and validity \cite{Baden2021ThreeGI}.
Our work follows previous research that aims to produce more interpretable topics in natural language \cite{Mei2007AutomaticLO, lau2011automatic, wan-wang-2016-automatic}. Additionally, our design builds on seeded/anchored topic models \cite{andrzejewski_latent_2009, jagarlamudi_incorporating_2012, gallagher_anchored_2017, meng_discriminative_2020}, hierarchical topic models \cite{griffiths2003hierarchical, teh_hierarchical_2006, mimno2007mixtures, paisley2014nested}, as well as those that impose constraints on topics \cite{wallach2009rethinking, hu2014interactive}.

\paragraph{LLM-based content analysis:}
LLMs such as ChatGPT have enabled new prompting and embedding-based approaches to analyzing text.
Researchers have used prompting techniques on these LLMs for related content analysis tasks, including text clustering \cite{viswanathan2023large, zhang_clusterllm_2023, hoyle2023natural}, abstractive summarization \cite{liu2023abstractive}, and deductive qualitative coding \cite{tai2023use, chew2023llmassisted}. Prior work has explored topic modeling with contextualized embeddings from pre-trained models \cite{sia-etal-2020-tired, thompson_topic_2020, bianchi2021pretraining, grootendorst2022bertopic}. Recent work has also used LLMs to label \cite{rijcken_towards_2023} and evaluate topics \cite{stammbach2023revisiting} that are produced by an existing topic model.
\paragraph{Comparison to GoalEx:} 
\frameworkname\ most closely resembles GoalEx \cite{goalex}, but is tailored specifically for topic modeling. GoalEx focuses on clustering and corpus \emph{partitioning} but not overall topic set organization, which is a fundamental need for content analysis. Specifically, when refining topics, GoalEx only retains clusters such that each document is supported approximately once, which may remove topics that often appear together but have different meanings. Instead, our framework refines topics based on their semantics to retain coherent, informative topics. Additionally, where GoalEx assigns topics to documents individually, which scales poorly, \frameworkname\ simultaneously provides all prompt topics to each document for efficient scaling. While GoalEx was evaluated only on cluster recovery, \frameworkname~is benchmarked on stability and alignment with ground-truth topics, demonstrating its usefulness as a content analysis tool beyond text clustering.

%% file: sections/3-Framework.tex
\section{Methodology}
\frameworkname~consists of two main stages: topic generation (\S\ref{subsec:proposer}) and topic assignment (\S\ref{subsec:assigner}). Figure \ref{fig:pipeline} provides an illustrative overview of our framework.

\subsection{Stage 1: Topic Generation}
\label{subsec:proposer}
Broadly, we prompt an LLM to generate a set of topics given an input dataset, and then we further refine these topics to remove infrequently used ones and merge duplicates. The output of this step can optionally be fed into \frameworkname\ again to generate fine-grained subtopics (Appendix \ref{appendix:hierarchy}).
\paragraph{Generating new topics:}
Given a document $d$ from the corpus and a set of example topics $S$, the model is instructed to either assign $d$ to an existing topic in $S$ or generate a new topic that better describes $d$ and add it to $S$. We define a \emph{topic} to be a concise label paired with a broad one-sentence description, as in
\begin{quote}
\textbf{Trade}: Mentions the exchange of capital, goods, and services
\end{quote}
where ``Trade'' serves as the topic label.
Initially, S consists of a few human-written topics (our experiments use 2 example topics). These topics serve as ``few-shot''  demonstrations of the topic generation format. \textbf{Importantly, they do not need to be dataset-specific or representative of all possible topics in the corpus, as we show in Appendix \ref{appendix:example}}. This iterative process encourages newly generated topics to be distinctive and match the specificity seen in other topics. Notably, instead of running topic generation over the entire corpus, which can be extremely costly, we apply the process to a carefully constructed sample from the dataset (\S \ref{subsec:train-sample}).

\paragraph{Refining generated topics:}
Optionally, we can further refine the generated topics to ensure that the final topic list is coherent and non-redundant. 
We first use Sentence-Transformer embeddings \cite{reimers-gurevych-2019-sentence} to identify pairs of topics with cosine similarity $\geq$ 0.5.
We then prompt the LLM with five such topic pairs, instructing it to merge near-duplicate pairs where appropriate.
To address any minor topics that may have been overlooked in the prior step, we eliminate topics with low frequency of occurrence. To do this, we keep track of how frequently each topic gets generated. If a topic occurs below a ``removal'' threshold frequency,\footnote{We recommend trying different thresholds to make sure that important topics are not removed from the final list.} we consider that topic to be minor and remove it from the final list.


\subsection{Stage 2: Topic Assignment}
\label{subsec:assigner}
In the assignment stage, we aim to establish a valid and interpretable association between the generated topics and the documents in our datasets. We provide the LLM with our generated topic list, 2-3 examples, and a document, the topic(s) of which we are interested in obtaining. We then instruct the model to assign one or more topics to the given document. The final output contains the assigned topic label, a document-specific topic description, and a quote from the document that supports this assignment. The quoted text improves the verifiability of \frameworkname's assignments, which has been a long-standing concern with traditional methods such as LDA. A sample topic assignment is 
\begin{quote}
\textbf{Agriculture}: Mentions changes in agricultural export requirements (``...repeal of the agricultural export requirements...'')
\end{quote} where ``Agriculture'' is the assigned topic label, followed by the topic description and a quote from the document enclosed within parentheses.

\paragraph{Self-correction:}
\label{subsub:correction}
To address topic assignments with incorrect formatting or low quality, we incorporate a self-correction step \cite{shinn2023reflexion, sun2023pearl}. Specifically, we implement a parser to identify hallucinated or invalid assignments (e.g. ``None''/``Error''). Next, we provide the LLM with the identified documents along with the error type and prompt the model to reassign a valid topic. 

%% file: sections/4-Experiments.tex
\section{Experiments}
Our goal is to assess whether \frameworkname's outputs align with human-coded ground truth topics, as well as to test its robustness to various settings. Here, we describe our datasets, baseline methods, model configurations, and evaluation metrics. 
\subsection{Datasets}
\label{subsec:data}
We use two English-language datasets for evaluation: \texttt{Wiki} and \texttt{Bills}. See Appendix \ref{appendix:memorization} for statistics and discussion about data memorization.

\vspace{0.2cm}

\noindent\textbf{\texttt{Wiki}} \cite{MerityWiki} has 14,290 Wikipedia articles that meet a core set of editorial standards.
This dataset comes with 15 high-level, 45 mid-level, and 279 low-level human-annotated labels. 
\vspace{0.2cm}

\noindent\textbf{\texttt{Bills}} (\citealt{adler2018congressional}, compiled by \citealt{hoyle-etal-2022-neural}) contains 32,661 bill summaries from the $110-114^{\text{th}}$ U.S. congresses, with 21 high-level and 114 low-level human-annotated labels. 

\subsection{Baselines}
\label{sec:baselines}
We consider three popular topic models that follow different paradigms: LDA, BERTopic, and SeededLDA. We control the number of topics $k$ to be equal to the number of topics generated by \frameworkname\ for a fair comparison. We adopt~\citet{hoyle_is_2021}'s data preprocessing pipeline for LDA and SeededLDA. We do not preprocess for BERTopic.\footnote{\href{https://maartengr.github.io/BERTopic/faq.html\#should-i-preprocess-the-data}{https://maartengr.github.io/BERTopic/faq.html\#should-i-preprocess-the-data}}

\paragraph{LDA:} We use the MALLET \cite{McCallumMALLET} implementation of LDA with Gibbs sampling \cite{griffiths2002gibbs},
which boasts strong alignment with human codes \cite{srivastava2017autoencoding, hoyle_is_2021} and high stability compared to neural topic models \cite{hoyle-etal-2022-neural}. We set $|V|$ = 15,000, $\alpha = 1.0$, $\beta = 0.1$, and run LDA for 2,000 iterations with optimization at every 10 intervals.

\paragraph{BERTopic:} BERTopic is a popular neural topic model that obtains topics by performing clustering on Sentence-Transformer embeddings of documents \cite{grootendorst2022bertopic}. We maintain all default hyperparameters.

\paragraph{SeededLDA:} SeededLDA \cite{jagarlamudi_incorporating_2012} incorporates seed topics to steer the resulting word-topic and document-topic distributions towards topics of interest to the users. We maintain all default hyperparameters 

\subsection{Sampling documents for \frameworkname}
\label{subsec:train-sample}
The number of documents used during the topic generation phase is a critical parameter of \frameworkname. Given enough time and money (if using closed-source LLM APIs), we could use the entire training corpus for topic generation. 
However, this is impractical given its high cost and also unnecessary as confirmed in practice below. We thus sample a document subset uniformly at random for topic generation, which results in 1,000 documents from \texttt{Bills} and 1,100 documents from \texttt{Wiki}.

\paragraph{How many documents should we sample?}
We recommend that users either choose a sample size that fits their budget or run topic generation incrementally and stop when no new topics are generated for some threshold (e.g., 200 documents). To assess the topic coverage of this approach, we examine the topics generated after reaching this threshold and check whether our refinement process removed any of them. In both datasets, the number of new topics that remain after refinement plateaus after this threshold - after around 600 documents have been generated. 

\subsection{\frameworkname\ implementation details}
\label{subsub:default}
Our default \frameworkname\ setting uses OpenAI's GPT-4 to generate topics and GPT-3.5-turbo to assign topics to documents.\footnote{Results in the default settings were finalized in August 2023. Ablation results were finalized in October 2023.} To encourage deterministic outputs,\footnote{We note that OpenAI's LLMs possess some degree of non-determinism even after fixing the decoding hyperparameters to perform greedy decoding.} we set \texttt{max\_tokens} to 300 and \texttt{temperature} and \texttt{top\_p} to 0. We truncate longer documents to fit within the context window size of LLMs.\footnote{The context window size for GPT-4 is 8,192 tokens, whereas GPT-3.5-turbo has a context window of 4,096 tokens.}
Since \texttt{Bills} and \texttt{Wiki} have one-to-one mappings between documents and labels, we modify the assigner prompt to assign only one topic per document, although we emphasize that the method does not require a single assignment. We set the removal frequency threshold to 10 and 5 for \texttt{Bills} and \texttt{Wiki}, respectively. We enable self-correction with a retry limit of 10 and end up resolving all topic hallucinations and formatting issues within this limit. For evaluation purposes, we sample 8,024 documents from \texttt{Wiki} and 15,242 documents from \texttt{Bills} that are not included in the topic generation sample.

\subsection{Evaluation Setup}
To assess the usefulness of \frameworkname~as an automated content analysis tool, we evaluate the topic alignment and stability of the generated topics, following~\citet{hoyle-etal-2022-neural}.
\subsubsection{Topical alignment}
Since \frameworkname\ is not a probabilistic model like LDA, we compare it to baselines by assessing the alignment between predicted topic assignments and topic labels in the ground truth, following prior work~\cite{pmlr-v28-chuang13, poursabzi2016alto, korencic_topic_2021, hoyle-etal-2022-neural}. For other methods, we assign each document to its most probable topic.
Given a set of ground-truth classes and a set of predicted assignment clusters, we assessed alignment between the two sets using three external clustering metrics.
\paragraph{Purity.} We use the harmonic mean of purity \cite{zhao2001criterion} and inverse purity to match each ground-truth category with the cluster that has the highest combined precision and recall \cite{amigo2009comparison}. Purity yields a score of close to 0 for random assignments and nearly 1 for strongly consistent assignments. 

\paragraph{Adjusted Rand Index.} The Rand Index measures the pairwise agreement between two sets of clusters. Adjusted Rand Index (ARI) further corrects for chance, yielding a score near 0 for random assignments and near 1 for consistent assignments \cite{hubert1985comparing, vinh2009information}.

\paragraph{Normalized Mutual Information.} Mutual Information (MI) measures the amount of shared information between two sets of clusters \cite{shannon1948mathematical}. NMI normalizes MI to a value between 0 and 1, making MI less sensitive to a varying number of clusters \cite{strehl2002cluster}.

\paragraph{Comparison of metrics:} $P_1$, ARI, and NMI provide complementary perspectives through set matching, counting pairs, and variation of information \cite{meilua2007comparing}. 
$P_1$ emphasizes cluster purity, not so much the distribution of ground-truth labels. ARI, unlike NMI, is adjusted for chance, but it does not account for class imbalance, making it more sensitive to the number of predicted clusters. 

\subsubsection{Stability}
We also assess the robustness of \frameworkname~to changes in prompts and corpus samples for generation by measuring whether \frameworkname\ maintains high topical alignment in these modified settings.

\paragraph{Out-of-domain prompts.} In \frameworkname, users can change the example topics and few-shot examples in the prompts to tailor the method to their dataset. We explore whether prompts written for one dataset can work on another by applying prompts for \texttt{Wiki} on the \texttt{Bills} dataset. 

\paragraph{Additional example topics.} We assess the impacts of additional example topics on \frameworkname's performance in \texttt{Bills}. Our original prompt for topic generation has two example topics. We expand the prompt by adding three more topics. 

\paragraph{Shuffling sampled documents for topic generation.} We shuffle documents in the generation sample of \texttt{Bills} to understand the importance of the order in which LLM processes documents.

\paragraph{Using a different sample for topic generation.} To evaluate \frameworkname's robustness to data shift, we apply \frameworkname~to a different generation sample from \texttt{Bills} and examine the results' variation.

%% file: sections/5-Results.tex
\section{Results}
Our results demonstrate \frameworkname's strong topical alignment with ground truth and robustness to variations in prompts and data. Our human evaluation also show that the \emph{semantic} content of \frameworkname's generated topics (ignoring document assignment) is significantly more aligned with ground-truth topics on both datasets. Furthermore, we explore the use of an open-source model for topic generation.
\begin{table*}
\small
\centering
\begin{tabular}{p{0.75cm}p{4.5cm}p{0.4cm}p{0.4cm}p{0.4cm}p{0.4cm}p{0.4cm}p{0.4cm}p{0.4cm}p{0.4cm}p{0.4cm}p{0.4cm}p{0.4cm}p{0.4cm}}
    \toprule
    \multirow{2}{*}{Dataset} & \multirow{2}{*}{Setting} & \multicolumn{3}{c}{TopicGPT} & \multicolumn{3}{c}{LDA} & \multicolumn{3}{c}{BERTopic} & \multicolumn{3}{c}{SeededLDA}\\
    \cmidrule(lr){3-5}\cmidrule(lr){6-8}\cmidrule(lr){9-11}\cmidrule(lr){12-14}
    & &$P_1$&ARI&NMI&$P_1$&ARI&NMI&$P_1$&ARI&NMI&$P_1$&ARI&NMI\\
    \midrule
    \multirow{2}{*}{\texttt{Wiki}} & Default setting ($k$=31) & \textbf{0.73} & \textbf{0.58} & \textbf{0.71} & 0.59 & 0.44 & 0.65 & 0.54 & 0.24 & 0.50 & 0.61 & 0.47 & 0.65\\
    & Refined topics ($k$=22) & \textbf{0.74} & \textbf{0.60} & \textbf{0.70} & 0.64 & 0.52 & 0.67 & 0.58 & 0.28 & 0.50 & 0.62 & 0.51 & 0.65\\
    \midrule
    \multirow{2}{*}{\texttt{Bills}} & Default setting ($k$=79) & \textbf{0.57} & \textbf{0.42} & \textbf{0.52} & 0.39 & 0.21 & 0.47 & 0.42 & 0.10 & 0.40 & 0.50 & 0.28 & 0.43\\
     & Refined topics ($k$=24) & \textbf{0.57} & \textbf{0.40} & \textbf{0.49} & 0.52 & 0.32 & 0.46 & 0.39 & 0.12 & 0.34 & 0.52 & 0.31 & 0.45\\\midrule
     \multicolumn{14}{c}{\scriptsize{\emph{\frameworkname\ stability ablations, baselines controlled to have the same number of topics (\textbf{$k$}).}}}\vspace{0.1cm} \\
    \multirow{5}{*}{\texttt{Bills}} & Different generation sample ($k$=73) & \textbf{0.57} & \textbf{0.40} & \textbf{0.51} & 0.41 & 0.23 & 0.47 & 0.38 & 0.08 &  0.38 & 0.40 & 0.21 & 0.44\\
    & Out-of-domain prompts ($k$=147) & \textbf{0.55} & \textbf{0.39} & \textbf{0.51} & 0.31 & 0.14 & 0.47 & 0.35 & 0.07 & 0.41 & 0.29 & 0.13 & 0.44\\
    & Additional example topics ($k$=123) & \textbf{0.50} & \textbf{0.33} & \textbf{0.49} & 0.33 & 0.15 & 0.46 & 0.36 & 0.07 & 0.40 & 0.33 & 0.15 & 0.44 \\
    & Shuffled generation sample ($k$=118) & \textbf{0.55} & \textbf{0.40} & \textbf{0.52} & 0.33 & 0.16 & 0.47 & 0.36 &  0.08 & 0.40 & 0.34 & 0.18 & 0.44\\
    & Assigning with Mistral ($k$=79) & \textbf{0.51} & \textbf{0.37} & 0.46 & 0.39 & 0.21 & \textbf{0.47} & 0.42 & 0.10 & 0.40 & 0.50 & 0.28 & 0.43\\
    \bottomrule
\end{tabular}
\caption{\label{tab:result}Topical alignment between ground-truth labels and predicted assignments. Overall, \frameworkname\ achieves the best performance across all settings and metrics compared to LDA, BERTopic, and SeededLDA. The number of topics used in each setting is specified as $k$. The largest values in each metric and setting are \textbf{bolded}.}
\end{table*}

\subsection{\frameworkname~is strongly aligned to ground truth labels}
Our experiments (Table \ref{tab:result}) show that \frameworkname~identifies topics that are substantially more aligned with human-annotated labels than baselines, and that this improvement holds across all datasets, settings, and metrics. Of the three baselines, LDA is generally comparable to SeededLDA but superior to BERTopic for all metrics, suggesting that LDA remains a strong baseline.
However, none of the baselines approaches the performance of \frameworkname: for example, \frameworkname\ achieves post-refinement harmonic purity scores $P_1$ of 0.74 and 0.57 on \texttt{Wiki} and \texttt{Bills}, respectively, compared to 0.64 and 0.52 for LDA, 0.58 and 0.39 for BERTopic, and 0.62 and 0.52 for SeededLDA.

\paragraph{Where does \frameworkname\ disagree with the ground truth?} To fully understand the disagreement between \frameworkname~and human labels, we examine five assignments where the ground truth topics and \frameworkname's assignments in the default setting do not match, and we find that each sampled document could reasonably be assigned multiple topics.
Therefore, \frameworkname's assignments can still be valid even if they differ from the ground truth labels. For example, the second document in Table \ref{tab:result-error-analysis} could fit either the ``\emph{Labor}'' or ``\emph{Transportation Safety}'' topics, though ``\emph{Labor}'' is more prominent.
If we allow \frameworkname~to assign multiple topics per document, can it successfully retrieve all appropriate topics? To check, we re-run topic assignment using a prompt that allows the assignment of multiple topics per document. With this updated prompt, three out of the five \texttt{Wiki} examples are assigned to the ground truth as well as the originally assigned topic (Table \ref{tab:result-error-analysis}). 
Thus, we recommend that practitioners use multi-label assignment prompts to extract as many relevant topics as possible. 

\subsection{\frameworkname~is stable}
Broadly speaking, \frameworkname~produces comparable topical alignment with the ground truth across all modified experimental settings for the \texttt{Bills} data (lower portion of Table \ref{tab:result}). The setting with additional example topics obtains the worst performance, suggesting that simply adding more example topics is not always helpful. 
Too many example topics may overwhelm the model and lead to poorer coherence, as the model tries to fit diverse topics rather than consolidating around the most salient themes.
We recommend keeping the example topic list small (2-3 high-quality topics) for best results, rather than lengthening the list arbitrarily. 

\paragraph{Consistency between multiple settings of \frameworkname:}To further evaluate the consistency of \frameworkname's topic assignments, we computed alignment scores between the default setting and each modified setting. As a benchmark, we ran LDA 10 times with $k$=79 topics and calculated the average internal alignment between each pair of runs. Table \ref{tab:stability-comparison} shows \frameworkname's assignments were highly stable across settings, with all metrics within a tight 0.05 range. \frameworkname~demonstrated greater stability than LDA in terms of $P_1$ and ARI, while achieving comparable NMI. Interestingly, \frameworkname~produces slightly different outputs between two runs with identical settings, a possible result of adding randomness to the self-correction process (\S\ref{subsub:correction}) as well as LLM API non-determinism.

\begin{table*}[t!]
\small
\centering
\begin{tabular}{llccccccccc}
    \toprule
    Method & Setting & $P_1$ & ARI & NMI \\
    \midrule 
    LDA & Default setting ($k$=79) & 0.64 & 0.55 & 0.71\\
    \midrule
    \multirow{5}{*}{TopicGPT} & Different generation sample ($k$=73) & 0.67 & 0.61 & 0.69\\
    & Out-of-domain prompts ($k$=147) & 0.69 & 0.63 & 0.69 \\
    & Additional example topics ($k$=123) & 0.69 & 0.59 & 0.70 \\
    & Shuffled generation sample ($k$=118) & 0.70 & 0.63 & 0.70\\
    & Running the pipeline twice ($k$=79) & 0.95 & 0.92 & 0.92\\
    \bottomrule
\end{tabular}
\caption{\label{tab:stability-comparison}Stability of topic assignments of \frameworkname~and LDA in the \texttt{Bills} dataset, as measured by the topical alignment between topic assignments of each modified setting against the default setting (unlike Table~\ref{tab:result} which reports alignment against \emph{ground-truth} assignments).} 
\end{table*}
\input{sections/5a-Human}
See Appendix \ref{appendix:rtl} for an additional qualitative evaluation of TopicGPT in a different domain, compared with expert-derived qualitative categories. 
\begin{table*}[t!]
\scriptsize
\begin{tabular}{@{}p{0.75cm}p{6.5cm}p{1.5cm}p{3.0cm}p{2.5cm}@{}}
\toprule
  \textbf{Data} &
  \textbf{Document}&
  \textbf{Ground truth} &
  \textbf{\frameworkname\ assignment} &
  \textbf{LDA assignment} \\ \midrule
\texttt{Wiki} &
  \href{https://en.wikipedia.org/wiki/Grant_Park_Music_Festival}{Grant Park Music Festival} = The Grant Park Music Festival ( formerly Grant Park Concerts ) is an annual ten-week classical music concert series held in Chicago, Illinois, USA. It features the Grant Park Symphony Orchestra and Grant Park Chorus along with featured guest performers and conductors. The Festival has earned non-profit organization status. It claims to be the nation's only free, outdoor classical music series. The Grant Park Music Festival has been a Chicago tradition since 1931 when Chicago Mayor Anton Cermak suggested free concerts to lift the spirits of\dots &
  \textbf{Music} &
  \textbf{Music \& Performing Arts}: Discuss creation, production, and performance of music, as well as related arts and cultural aspects. & \textcolor{red}{\textbf{City infrastructure}}: city, building, area, new, park  \\
  \midrule
\texttt{Bills} &
  \href{https://www.govinfo.gov/bulkdata/BILLSUM/114/s/BILLSUM-114s2718.xml}{Perkins Fund for Equity and Excellence}.
  This bill amends the Carl D. Perkins Career and Technical Education Act of 2006 to replace the existing Tech Prep program with a new competitive grant program to support career and technical education. Under the program, local educational agencies and their partners may apply for grant funding to support: career and technical education programs that are aligned with postsecondary education programs, dual or concurrent enrollment programs and early college programs, certain evidence-based strategies and delivery models related to career and technical education, teacher and leader experiential \dots &
  \textbf{Education} &
  \textbf{Education}: Mentions policies and programs related to higher education and student loans. & \textcolor{red}{\textbf{Programs and grants:}} program, grants, grant, programs, state \\ \bottomrule
\end{tabular}
\caption{\label{tab:example}Example topic assignments from \frameworkname\ and LDA (showing top 5 topic words) on two documents. While \frameworkname's topics closely align with the ground truth, LDA's topics are influenced by frequently occurring words, causing it to overlook the overarching theme of the document. \frameworkname's topic labels and descriptions are both automatically generated, while LDA produces a bag of words that needs to be \textcolor{red}{manually labeled}.}
\end{table*}

\subsection{Implementing \frameworkname\ with open-source LLMs }
\label{sub:mistral}
We examine alternate LLMs for both topic assignment and topic generation and discover that while topic assignment can be feasibly performed with open-source LLMs, topic generation is too complex for all LLMs we tried other than GPT-4. 

\paragraph{Mistral-7B-Instruct for topic assignment:} 
We experiment with Mistral-7B-Instruct \cite{jiang2023mistral} for topic assignment to assess its feasibility as a lower-cost alternative to GPT-3.5-turbo. Mistral's topic assignments align reasonably well with human ground truth, though not to the same degree as GPT-3.5-turbo; the bottom row of Table \ref{tab:result} shows an absolute purity decrease of about $\approx 6$ points. However, the resulting assignments still outperform baseline methods on $P_1$ and ARI.

\paragraph{Mistral-7B-Instruct for topic generation:} We further test both Mistral and GPT-3.5-turbo as topic generation models, finding that both models struggle to follow formatting instructions for topic generation. In total, Mistral and GPT-3.5-turbo produced 1,418 and 151 topics, respectively. The large number of topics made it impossible to include all of them in a single topic assignment prompt. Additionally, most of the generated topics are overly specific with a low frequency of occurrence, meaning they would likely be removed during refinement. 
We emphasize that the instructions for topic generation are complex with many criteria (Table~\ref{tab:prompt-proposer-1}). As such, we recommend sticking with GPT-4 or models with similar capabilities to generate topics.

%% file: sections/5a-Human.tex
\subsection{\frameworkname\ topics are semantically close to ground truth}
\label{sec:human-eval}
The metrics reported above do not capture whether the topics are semantically aligned with ground truth. To this end, we qualitatively compare LDA outputs and \frameworkname's unrefined and refined topic generations and analyze the proportion of misaligned topics produced by each method.

\paragraph{Manual topic matching process:} Three annotators (the first author and two external annotators)\footnote{External annotators were recruited from our network and provided uncompensated annotations to support this research.}
went through the list of generated topics and assigned each topic to a ground truth class. If an exact match was not possible, the annotators labeled the generated topic as one of the three misaligned categories as below \cite{pmlr-v28-chuang13}:
\begin{enumerate}
    \item Out-of-scope: topics that are too narrow/broad compared to the associated ground truth.
    \item Missing topics: topics present in the ground truth but not in the generated outputs.
    \item Repeated topics: topics that are duplicates of other topics.
\end{enumerate}  
After completing the mapping, the percentage of misaligned topics was calculated for each annotator's mappings (Table \ref{tab:qual}). Each person completed a total of 6 mappings, including LDA outputs, unrefined and refined topic list for both datasets. Mapping instructions and interface can be found in Appendix \ref{appendix:annotation}.

\begin{table*}[t!]
\small
\centering
\begin{tabular}{llcccccccccccc}
    \toprule
    Dataset & Setting & Out-of-scope & Missing & Repeated & Total\\
    \midrule
    \multirow{3}{*}{\texttt{Wiki}} & LDA ($k$=31) & 46.3 & 4.3 & 11.9 & 62.4\\
    & Unrefined ($k$=31) & 38.7 & \textbf{0.0} & 1.1 & 39.8\\
    & Refined ($k$=22) & \textbf{30.3} & \textbf{0.0} & \textbf{0.0} & \textbf{30.3}\\
    \midrule
    \multirow{3}{*}{\texttt{Bills}} & LDA ($k$=79) & 56.1 & 2.1	& 22.0 & 80.2\\
    & Unrefined ($k$=79) & 65.0 & \textbf{1.3} & 3.8 & 70.1\\
    & Refined ($k$=24) & \textbf{27.8} & 4.2 & \textbf{0.0} & \textbf{31.9}\\
    \bottomrule
\end{tabular}
\caption{\label{tab:qual} Comparison of misaligned topic proportions between LDA and \frameworkname\ outputs. Values are averaged over three annotations and rounded to one decimal place. Across both datasets, \frameworkname\ (with refinement) achieves the lowest proportion of misaligned topics The best (lowest) misalignment proportion for each dataset is \textbf{bolded}.}
\end{table*}

\paragraph{\frameworkname\ contains far fewer misaligned topics than LDA, especially after refinement.} Compared to LDA, both \frameworkname's unrefined and refined topics are less likely to be misaligned overall (62.4\% for LDA vs. 38.7\% unrefined and 30.3\% refined). Annotators also noted that \frameworkname's outputs are much easier to work with compared to the ambiguous LDA outputs (see Table \ref{tab:example} for examples of LDA and \frameworkname's outputs). We notice that refinement consistently reduces the number of out-of-scope and repeated topics. On the other hand, refinement does increase the number of missing topics by 1 in the Bills dataset; upon closer examination, this missing topic was ``\emph{Culture}'', which appears infrequently in the Bills corpus (only 23 documents out of 32,661 documents). This might be acceptable depending on the use case, and we emphasize again that practitioners should try different refinement thresholds to avoid filtering out topics that are important to their research question.

%% file: sections/6-Discussion.tex
\section{Future Work}
\frameworkname\ is designed for inductive content analysis, assuming the user has some prior familiarity with the dataset's content. However, the framework can also be applied to data exploration scenarios where the user is unfamiliar with the dataset and employs topic models to gain insights. Future work could explore the potential of zero-shot prompting (without providing example topics or documents) with \frameworkname\ for such exploratory use cases.

Appendix \ref{appendix:hierarchy} presents a hierarchical extension of this framework. Evaluating the performance of this extension against established hierarchical topic models through rigorous comparative studies will be a valuable next step.



%% file: sections/7-Conclusion.tex
\section{Conclusion}
We introduce a prompt-based framework, \frameworkname, specifically designed for topic modeling. \frameworkname~addresses traditional topic models' interpretability and adaptability limitations by generating high-quality and descriptive topics. Our results demonstrate that \frameworkname~outperforms baseline topic models in terms of topic alignment with ground truth labels while also showing robustness across different prompts and data subsets. We release our pipeline so that interested researchers and practitioners can try our framework.

%% file: sections/Limitation.tex
\section*{Limitations}
\paragraph{Transparency concerns of closed-source models.} \frameworkname\ achieves optimal performance using GPT-4 for topic generation and GPT-3.5-turbo for assignment, both of which are closed-source LLMs. Unfortunately, we have limited transparency into their pre-training and instruction tuning datasets, as well as their architectural details. Future work can explore using a stronger open-source model for topic generation or fine-tuning an LLM for topic assignment. Our reliance on closed-source LLMs for topic generation reflects the current imbalance between the instruction-following abilities of closed and open-source models rather than a permanent limitation. We hope this will eventually be addressed by the rapid advances in open-source LLMs. 

\paragraph{Cost concerns of closed-source models.} \frameworkname's use of closed-source models also incurs costs for each run (see Table \ref{tab:cost} for our expenses). In addition to exploring open-source alternatives to these models, users can take advantage of the framework's modular design and tailor it to their use cases. For instance, if users are solely interested in obtaining a list of topics associated with a dataset, they can skip topic refinement and assignment steps or reduce the size of the corpus subset used for topic generation, thus saving running time and money. 

\paragraph{Dealing with context limits.} Another limitation of our current approach is the need to truncate documents to fit \frameworkname's context length limit. By only providing partial documents, we lose potentially valuable context and risk misrepresenting the contents of full documents. While truncation was necessary in these initial experiments, we recognize it is not an ideal solution. Future work can explore strategies to represent full documents within length limits, such as incrementally feeding in chunks of the document, sampling representative chunks, or providing a summarized version of the document. Additionally, future work can explore using long-context LLMs like GPT-4-turbo (128k tokens),\footnote{\href{https://platform.openai.com/docs/models/gpt-4-and-gpt-4-turbo}{https://platform.openai.com/docs/models/gpt-4-and-gpt-4-turbo}} Claude (200k tokens),\footnote{\href{https://www.anthropic.com/news/claude-3-family}{https://www.anthropic.com/news/claude-3-family}} or LLaMA-2-7B-32K (32K tokens)\footnote{\href{https://huggingface.co/togethercomputer/LLaMA-2-7B-32K}{https://huggingface.co/togethercomputer/LLaMA-2-7B-32K}} in the framework. 

\paragraph{Multilinguality.} We have not yet evaluated \frameworkname\ on non-English datasets. However, OpenAI's LLMs are pre-trained and instruction-tuned primarily on English language data, and the instruction-following capabilities of LLMs in non-English languages are thus notably degraded~\citep{huang2023not,li2023bactrianx}. We hope that future advances in multilingual LLMs will make \frameworkname\ more broadly accessible.

\section*{Ethical Considerations}
The risks posed by our framework are no greater than those inherent in the large language models that support it \cite{weidinger2021ethical}. Our human evaluation received approval from an institutional review board. All annotators (US-based) gave their informed consent and participated voluntarily, without compensation, to support our research. 

\section*{Acknowledgements}
We are grateful to Dzung Pham for the helpful discussion on evaluation metrics, Maria Antoniak, Brendan O'Connor, and Laure Thompson for their insights into topic model usage and evaluation, Chau-Anh and Upwork annotators for their assistance with the qualitative evaluation, and the UMass NLP community for helpful feedback throughout the project. This project was partially supported by awards IIS-2202506 and IIS-2046248 from the National Science Foundation (NSF).

%% file: sections/Appendix.tex
\onecolumn
\appendix
\label{sec:appendix}
\section{Hierarchical Implementation}
\label{appendix:hierarchy}
\frameworkname\ can be further extended to hierarchical topical modeling, enabling users to explore topics at various levels of granularity. Specifically, we treat the generated topics that remain after the refinement stage as \emph{top-level topics} and prompt the LLM for more specific subtopics at subsequent levels. We then provide the model with a \emph{topic branch} that contains a top-level topic $t$, a list of example subtopics $S'$, and the documents $d_t$ associated with the top-level topic $t$. With these inputs, we instruct the LLM to generate subtopics $t'$ that capture common themes among the provided documents. The model must also return specific documents supporting each subtopic to ensure subtopics are grounded in the documents rather than hallucinated. If the documents cannot fit into a single prompt, we divide them into different prompts and include subtopics generated by earlier prompts in subsequent ones. 

\begin{figure}[ht!]
\centerline{\includegraphics[width=0.5\columnwidth]{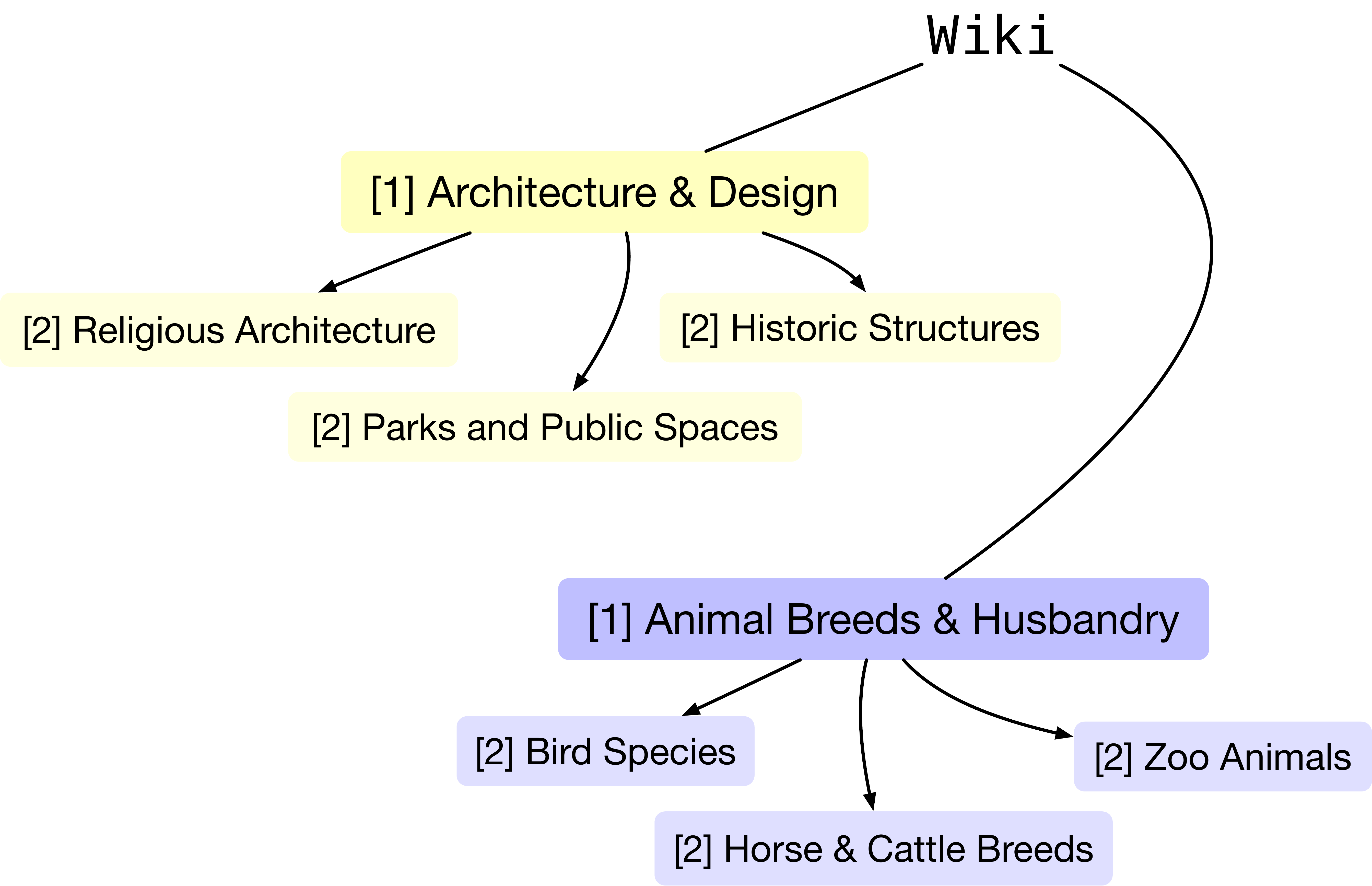}}
\caption{Example topic hierarchy for \texttt{Wiki}, with "Architecture \& Design" and "Animal Breeds \& Husbandry" as the top-level topics generated by \frameworkname. This hierarchical topic structure, in which the upper-level topics are broad enough to encompass more detailed subtopics, allows users to explore topics at different levels of specificity.}
\label{fig:wiki_hierarchy}
\end{figure}

We apply \frameworkname\ hierarchically to our datasets and find that the generated subtopics are informative and well-grounded in the documents associated with their parent topics. These subtopics successfully capture more narrow and nuanced themes within the broader parent topic, allowing for richer exploration and analysis compared to a flat topic list. Figure \ref{fig:wiki_hierarchy} provides a closer look into a portion of the generated topic hierarchy for the \texttt{Wiki} dataset. 
\newpage
\section{Comparison of \frameworkname~topics with human-curated qualitative categories in a different domain}
\label{appendix:rtl}
\input{sections/5b-Expert}

\newpage
\section{Probabilistic Justification for Data Sampling}
\label{appendix:sampling}
\begin{figure*}[ht!]
\centerline{\includegraphics[width=0.75\linewidth]{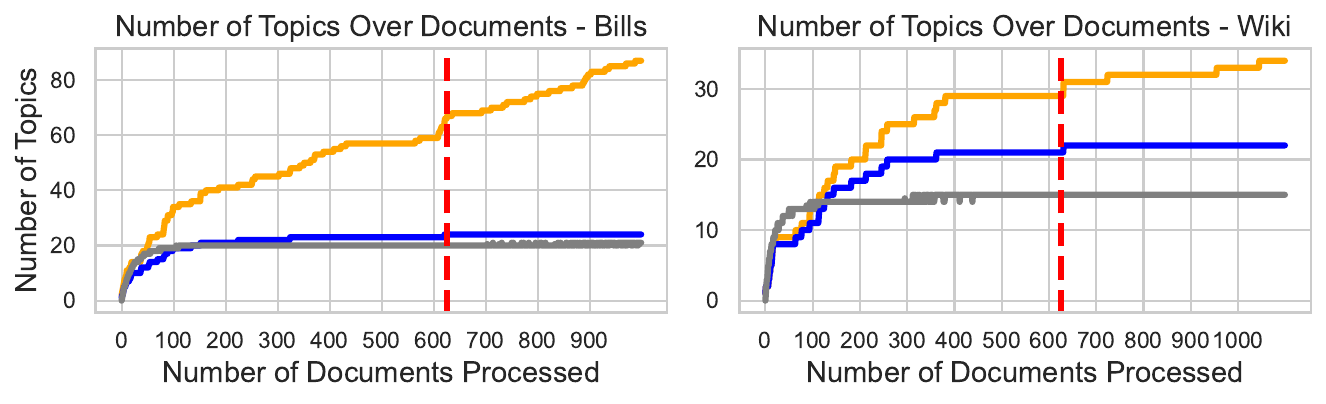}}
\caption{The number of topics generated over documents processed in the \texttt{Bills} and \texttt{Wiki} corpus. The \textcolor{gray}{grey} line indicates the number of expected topics, simulated using the empirical distribution of ground-truth topics for the datasets. For both datasets, we see a similar pattern - after a "topic drought" period marked by the dashed \textcolor{red}{red} line, the number of initially generated topics (\textcolor{orange}{orange} line) keeps increasing. However, the final refined topics (\textcolor{blue}{blue} line) and expected number of topics (\textcolor{gray}{grey} line) plateau, despite more documents being processed.}
\label{fig:wiki_train}
\end{figure*}

We plot the number of topics generated over documents processed in the \texttt{Bills} and \texttt{Wiki} corpus (Figure \ref{fig:wiki_train}). The graph confirms that after an initial "topic drought" period where few new topics are discovered, the number of initially generated topics continues increasing as more documents are processed. However, the final refined topics plateau over time, reaching a saturation point regardless of additional documents seen.

We present a probabilistic approach to sampling data for topic generation. The user sets a lower bound on the number of documents that should be assigned to the least-prevalent topic, $n_d$, which induces an upper bound on the number of topics, $K_u = \left \lfloor{N/n_d}\right \rfloor$. We can then model the sample of size $n_s$ as a draw from a uniform\footnote{If it were non-uniform, then we would have a topic with fewer than $n_d$ documents.} $\textbf{c} \sim \text{Multi}(n_s, K_u)$, where the expected number of zeros is then $\frac{(K_u - 1)^{n_s}}{K_u^{n_s-1}}$. A straightforward simulation can be used to find the $n_s$ that minimizes $p^*= \left| P\left( \min_k c_k = 0 \right) - \varepsilon \right|$, where $\varepsilon$ is a user-defined acceptable minimum probability of failing to find the least-prevalent topic. Taking the \texttt{Wiki} dataset, if we set $n_d=140$ (1\% of the corpus) then $n_s=1,100$ leads to $p^*\approx 0.005$.
\newpage
\section{Example Topics}
\label{appendix:example}
In all levels of topic generation, we include a list of example topics along with corresponding example documents. These example topics can be curated to inform the formatting and specificity of resulting topics. In this section, we offer suggestions on crafting example topics and provide corresponding illustrations of their impacts on the resulting topics.

\begin{enumerate}
    \item The format of the example topics should resemble the desirable output format, which contains a topic level and a concise topic label. \\

        \textit{Example 1: Using long example topic labels}\\
        \text{Example topics:}\\
        \text{[1] Trade Policies} \\
        \text{[1] Agricultural Policies}\\ 
        \newline
        \text{Generated topics (using GPT-4):}\\
        \text{[1] Social Security Policies}\\
        \text{[1] Food Safety Policies}\\

        \textit{Example 2: Using short example topic labels.}
        \text{Example topics:}\\
        \text{[1] Trade} \\
        \text{[1] Agriculture}\\ 
        \newline
        \text{Generated topics (using GPT-4):}\\
        \text{[1] Social Security}\\
        \text{[1] Food Safety}
    
    \item The example topics should match the degree of granularity you would expect in a given hierarchy level
    
        \textit{Example 1: Using abstract example topics for level 2.}\\
        \text{Example topics:}\\
        \text{[1] Music \& Performing Arts}
        \begin{itemize}
            \item\text{[2] Music}
            \item\text{[2] Performing Arts}
        \end{itemize}

        \text{Generated topics (using GPT-4):}\\
        \text{[1] Engineering \& Technology}
        \begin{itemize}
            \item\text{[2] Civil \& Transportation Engineering}
            \item \text{[2] Electrical Engineering}
        \end{itemize}

        \textit{Example 2: Using detailed example topic for level 2.}\\
        \text{Example topics:}\\
        \text{[1] Music \& Performing Arts}
        \begin{itemize}
            \item\text{[2] Albums}
            \item\text{[2] Songs}
        \end{itemize}

        \text{Generated topics (using GPT-4):}\\
        \text{[1] Engineering \& Technology}
        \begin{itemize}
            \item\text{[2] Road \& Highway Systems}
            \item\text{[2] Microwave Technology}
        \end{itemize}
    
\end{enumerate}
\newpage
\section{Varying number of topics $k$ for LDA baseline}
In the original experiments, we compare \frameworkname, which does not require $k$ to be specified, to baselines (LDA, BERTopic, SeededLDA) that do require k to be specified. We choose to set $k$ to the number of topics returned by \frameworkname\ to ensure a fair comparison with respect to our metrics. We further justify our decision with experiments that demonstrate \frameworkname’s superior topic alignment regardless of $k$ values. 

We compare \frameworkname\ in the default and refined settings to LDA across different values of k (15, 25, 50, 75, 100). As shown in Table \ref{tab:varying-k}, \frameworkname\ (with and without refinement) outperforms LDA on external clustering metrics in all settings. This experiment shows that regardless of k, \frameworkname\ has better alignment with ground-truth topics compared to LDA.

\begin{table}[ht]
\centering
\small
\begin{tabular}{@{}llccc@{}}
\toprule
Data             & Setting                                  & P1   & ARI  & NMI  \\ \midrule
\texttt{Bills} (k=21)    & \frameworkname\ (default setting, k=79)          & \textbf{0.57} & \textbf{0.42} & \textbf{0.52} \\
                 & \frameworkname\ (refined topics, k=24)          & 0.57 & 0.40 & 0.49 \\
                 & LDA (k=15)                               & 0.51 & 0.32 & 0.44 \\
                 & LDA (k=25)                               & 0.53 & 0.32 & 0.47 \\
                 & LDA (k=50)                               & 0.48 & 0.27 & 0.48 \\
                 & LDA (k=75)                               & 0.40 & 0.22 & 0.47 \\
                 & LDA (k=100)                              & 0.35 & 0.18 & 0.47 \\
\midrule
\texttt{Wiki} (k=15)     & \frameworkname\ (default setting, k=31)          & 0.73 & 0.58 & \textbf{0.71} \\
                 & \frameworkname\ (refined topics, k=22)          & \textbf{0.74} & \textbf{0.60} & 0.70 \\
                 & LDA (k=15)                               & 0.67 & 0.58 & 0.69 \\
                 & LDA (k=25)                               & 0.61 & 0.48 & 0.66 \\
                 & LDA (k=50)                               & 0.51 & 0.36 & 0.63 \\
                 & LDA (k=75)                               & 0.43 & 0.26 & 0.61 \\
                 & LDA (k=100)                              & 0.39 & 0.22 & 0.60 \\ \bottomrule
\end{tabular}
\caption{Performance of \frameworkname\ and LDA of various $k$ values. The best performance in each data and metric setting is highlighted.}
\label{tab:varying-k}
\end{table}

\newpage
\section{Human annotations on topical alignment between \frameworkname\ and ground truth of \texttt{Bills} dataset}
\label{appendix:annotation}
\begin{figure*}[ht!]
    \centering
    \includegraphics[width=1\linewidth]{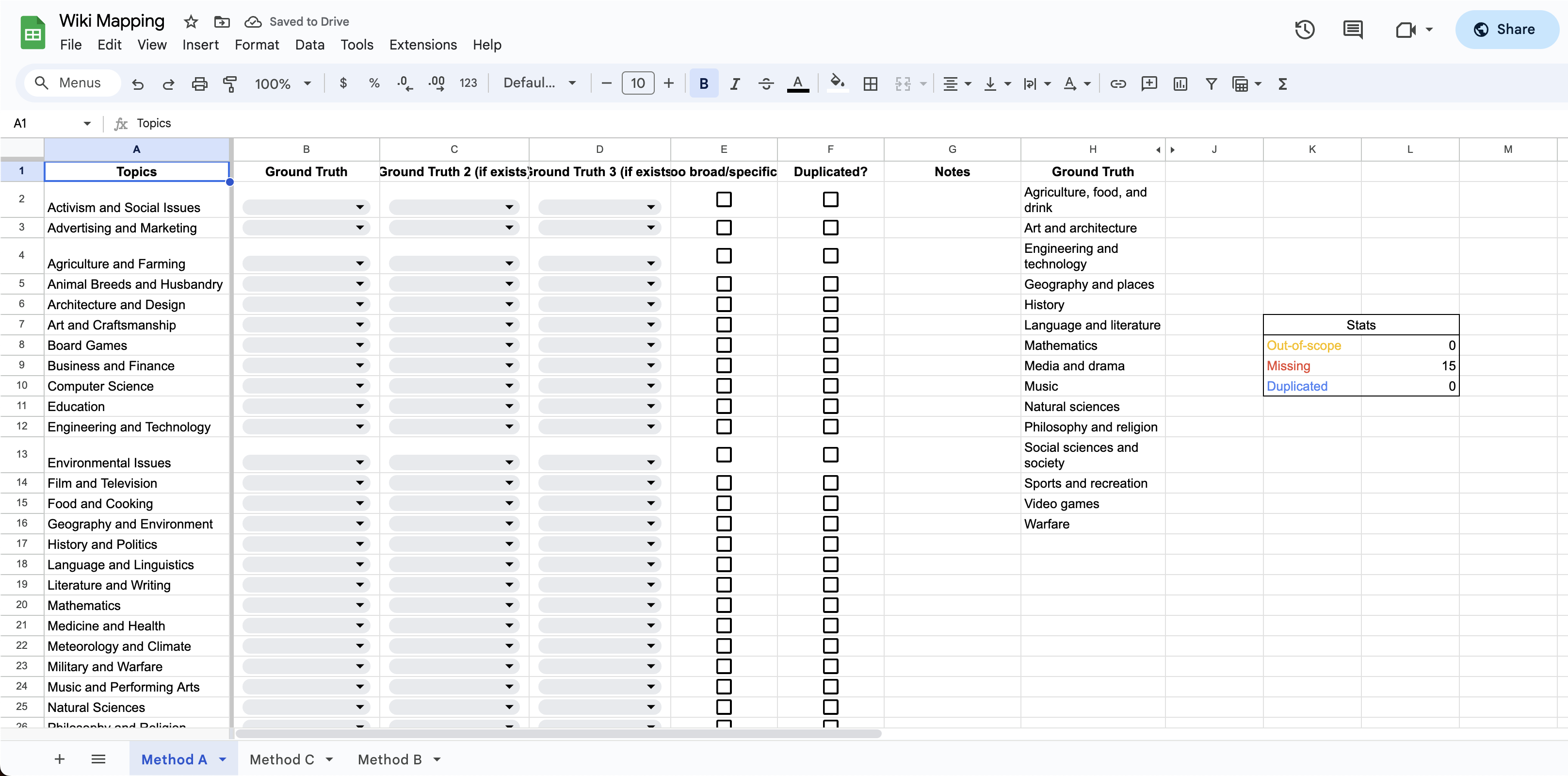}
    \caption{Mapping interface for human annotators. After reading the instructions in \ref{tab:instruction}, the annotators perform mapping between generated and ground-truth topics.}
    \label{fig:interface}
\end{figure*}

In this section, we provide additional details on our human evaluation (Section \ref{sec:human-eval}). Table \ref{tab:qual-full} shows the proportion of misaligned topics determined by each annotator. Table \ref{tab:instruction} and  Figure \ref{fig:interface} show the mapping and interface of the human annotation task.
\begin{table*}[ht]
\small
\begin{tabular}{llcccccccccccc}
    \toprule
    \multirow{2}{*}{Dataset} & \multirow{2}{*}{Setting} & \multicolumn{3}{c}{Out-of-scope} & \multicolumn{3}{c}{Missing} & \multicolumn{3}{c}{Repeated} & \multicolumn{3}{c}{Total}
    \\\cmidrule(lr){3-5}\cmidrule(lr){6-8}\cmidrule(lr){9-11}\cmidrule(lr){12-14}
    & & P1 & P2 & P3 & P1 & P2 & P3 & P1 & P2 & P3 & P1 & P2 & P3\\
    \midrule \midrule
    \multirow{3}{*}{\texttt{Wiki}} & LDA ($k=31$) & 74.2 & 19.4 & 45.2 & 3.2 & 6.5 & 3.2 & 9.7 & 6.5 & 19.4 & 87.1 & 32.4 & 67.8\\
    & Unrefined ($k=31$) & 67.7 & 25.8 & \textbf{22.7} & \textbf{0.0} & \textbf{0.0} & \textbf{0.0} & 3.2 & \textbf{0.0} & \textbf{0.0} & 70.9 & 25.8 & 22.7\\
    & Refined ($k=22$) & \textbf{59.1} & \textbf{9.1} & \textbf{22.7} & \textbf{0.0} & \textbf{0.0} & \textbf{0.0} & \textbf{0.0} & \textbf{0.0} & \textbf{0.0} & \textbf{59.1} & \textbf{9.1} & \textbf{22.7}\\
    \midrule
    \multirow{3}{*}{\texttt{Bills}} & LDA ($k=79$) & 64.6 & 50.6 & 53.2 & 2.5 & \textbf{2.5} & 1.3 & 16.5 & 22.8 & 26.6 & 83.6 & 75.9 & 81.0\\
    & Unrefined ($k=79$) & 72.2 & 62.0 & 60.8 & \textbf{1.3} & \textbf{2.5} & \textbf{0.0} & 6.3 & 2.5 & 2.5 & 79.8 & 67.1 & 63.3\\
    & Refined ($k=24$) & \textbf{45.8} & \textbf{4.2} & \textbf{33.3} & 4.2 & 4.2 & 4.2 & \textbf{0.0} & \textbf{0.0} & \textbf{0.0} & \textbf{50.0} & \textbf{8.3} & \textbf{37.5}\\
    \bottomrule
\end{tabular}
\caption{\label{tab:qual-full}Proportion of misaligned topics in LDA's outputs as well as \frameworkname's unrefined and refined outputs (rounded to one decimal place). The lowest values in each column are bolded.}
\end{table*}

\begin{table*}[ht]
\small
\begin{tabular}{p{\linewidth}}
\toprule
\textbf{Mapping Instruction for Human Annotators}
\\\midrule
\textbf{Introduction}
We ask you to compare categories created by two topic models with those created by humans for the same dataset. This comparison will help us understand whether these methods effectively cover relevant topics and accurately represent them. Additionally, we request that you provide a short discussion (1-2 paragraphs) for each method on how helpful and understandable the categories are for each of the two methods.\\

\textbf{Terminologies}
\begin{itemize}
    \item Out-of-scope categories: These are categories that are irrelevant compared to the true categories.  For example, “Albums” and “Songs” would be too narrow compared to a true category of “Music”.
    \item Repeated categories: These are categories that are unnecessary repeats or very similar to others. For example, “year month time ago day life” and “time hours month cycle year” are near duplicates.
    \item Missing categories: These are essential categories that appear in the true categories but are absent in the categories generated by the coding schemes.
\end{itemize}

\textbf{Instructions}

0. The sheet name (at the bottom) indicates the method used to get these categories.
Method A and C represents categories in natural language, with category labels and descriptions. 
Method B represents categories in terms of keywords, and you need to decide what the category is about. For example, the set of keywords “year month time ago day life” could be interpreted as referring to a " Time " category.  \\

1. For each category in the first column, choose 1, 2, or 3 corresponding true categories. For example, suppose Method A produced a category “Outdoor activities and sports”, and suppose two of the true categories were “Outdoor sports” and “Enjoying being outside” (not a real example). In this case you might choose to include both true categories. \\
If the category is too specific or too general compared to the associated true categories, check the box in the “Out-of-scope” column. 
If there are categories that are unnecessary repeats or very similar to others, only one instance of the category should be left alone and all the other instances should be marked as “Repeated”. 
The number of “Missing” will be automatically calculated (you will not be required to do anything about this). 

2. Provide a brief comment here (1-2 paragraphs) regarding the usability and interpretability of generated categories in corpus exploration and analysis. Here are some example questions to consider. Please note the specific method when making your comments.
\\\bottomrule
\end{tabular}
\caption{\label{tab:instruction} Instructions on the mapping process, provided in the form of a Google Doc for human annotators. }
\end{table*}
\newpage

\newpage
\section{Topical alignment metrics}
\label{appendix:metrics}
Here, we provide more details on the external clustering metrics used to measure topical alignment.
\paragraph{Purity.} Purity measures the degree to which predicted clusters contain data points predominantly from a single ground-truth class \cite{zhao2001criterion}. Purity is computed by assigning each predicted cluster to the ground-truth class that occurs most frequently within it, and then counting the number of correctly assigned documents and dividing by the number of clustered items $N$, 
\begin{equation}
\small
\mathrm{Purity}(\mathbf{\Omega}, \mathbf{C}) = \sum_k \frac{|\omega_k|}{N} \max_{j} \mathrm{Precision}(\omega_k, c_j)
\end{equation}
where the precision of a cluster $\omega_k$ for a given ground-truth class $\mathbf{C}_i$ is calculated as: 
\begin{equation}
\small
\mathrm{Precision}(\omega_k, c_j) = \frac{|\omega_k\cap c_j|}{|\omega_k|}
\end{equation}
While high purity indicates low intra-cluster noise, it does not reward grouping items from the same ground-truth class. Purity reaches its maximum value of 1 when each cluster contains just one item. Inverse Purity, defined as $\mathrm{Purity}^{-1} = \mathrm{Purity}(\mathbf{C}, \mathbf{\Omega})$, addresses this by rewarding clustering data points from the same class together. However, Inverse Purity fails to penalize mixing items from different ground-truth categories.~\citet{amigo2009comparison} show that  the harmonic mean of Purity and Inverse Purity balances these two objectives:
\begin{equation*}
\small
P_1 = \sum_k \frac{|c_k|}{N} \max_{j} {F(c_j, \omega_k)}
\end{equation*}
where 
{\small
\begin{equation}
F(c_k, \omega_j) = \frac{2*\mathrm{Precision}(c_k, \omega_j) * \mathrm{Recall}(c_k, \omega_j)}{\mathrm{Precision}(c_k, \omega_j) + \mathrm{Recall}(c_k, \omega_j)}
\end{equation}
}
and 
\begin{equation}
\small
    \mathrm{Recall}(\mathbf{C}, \mathbf{\Omega}) = \mathrm{Precision}(\mathbf{\Omega}, \mathbf{C})
\end{equation} 

\paragraph{Adjusted Rand Index.} The Rand Index measures the pairwise agreement between two sets of clusterings \cite{rand1971objective}:
\begin{equation}
\small
    RI(\mathbf{\Omega}, \mathbf{C}) = \frac{TP + TN}{{TP + FP + FN + TN}}
\end{equation} where $TP$ is the number of true positives, $TN$ is the number of true negatives, $FP$ is the number of false positives, and $FN$ is the number of false negatives.
\noindent Adjusted Rand Index (ARI) further corrects for chance by comparing it to the expected value
\begin{equation}
\small
    ARI(\mathbf{\Omega}, \mathbf{C}) = \frac{\text{RI} - \mathbb{E}[\text{RI}]}{\max(\text{RI}) - \mathbb{E}[\text{RI}]}
\end{equation} where $\mathbb{E}[\text{RI}]$ stands for expected rand index \cite{hubert1985comparing, vinh2009information}. ARI yields a score close to 0 for random cluster assignments and near 1 for strongly consistent assignments.

\paragraph{Normalized Mutual Information.} Mutual Information (MI) measures the amount of shared information between two sets of clusterings \cite{shannon1948mathematical}.
Normalized Mutual Information (NMI) normalizes the MI score to a value between 0 and 1, making MI less sensitive to a varying number of clusters \cite{strehl2002cluster}:
\begin{equation}
\small
    NMI(\mathbf{\Omega}, \mathbf{C}) = \frac{I(\mathbf{\Omega}, \mathbf{C})}{[H(\mathbf{\Omega}) + H(\mathbf{C})]/2}
\end{equation}
where $I(\mathbf{\Omega}, \mathbf{C})$ is the mutual information between $\Omega$ and $\mathbf{C}$, $H(\mathbf{\Omega})$ and $H(\mathbf{C})$ are the entropy of $\mathbf{\Omega}$ and $\mathbf{C}$.  

\newpage
\section{Experiment datasets}
Table \ref{tab:data-overview} shows high-level statistics for \texttt{Wiki} and \texttt{Bills} datasets. Table \ref{tab:cost} details the cost of running \frameworkname\ on each dataset. 
\begin{table}[ht]
\small
\centering
\begin{tabular}{p{0.5\linewidth}ll}
 \toprule
 \textbf{Stages} & \texttt{Bills} & \texttt{Wiki}\\
 \midrule
 Topic generation & \$30 & \$90 \\
 Topic refinement & \$10 & \$5\\
 Assignment + self-correction & \$48 & \$60 \\
 \midrule
 Total & \$88 & \$155\\
 \bottomrule
\end{tabular}
\caption{\label{tab:cost}Estimated cost (in US dollars) of running \frameworkname\ on  the \texttt{Bills} and \texttt{Wiki} datasets. Though smaller in size, the Wiki dataset incurred a much higher cost to run \frameworkname~than the \texttt{Bills} dataset due to its longer document length - on average 3,412 tokens/document compared to just 261 tokens/document in \texttt{Bills}.}
\end{table}

\label{appendix:memorization}
\begin{table}[ht]
\centering
\small
\begin{tabular}{lrr}
\toprule
& \texttt{Bills} & \texttt{Wiki}\\\midrule
\# docs & 32,661 & 14,290\\
avg. \# tokens / doc & 261 & 3,412\\
\# test examples & 15,242 & 8,024\\
\# topic generation docs & 1,000 & 1,100\\

\bottomrule
\end{tabular}
\caption[]{\label{tab:data-overview} Dataset statistics of \texttt{Bills} and \texttt{Wiki}. The average number of tokens per document was calculated using \texttt{tiktoken} BPE tokenizer.\footnotemark}
\end{table}
\footnotetext{\href{https://github.com/openai/tiktoken}{https://github.com/openai/tiktoken}}
\paragraph{Potential effects of data memorization on \frameworkname's performance}
\texttt{Bills} is not likely to appear in LLMs' training data because the label text is not directly associated with the corresponding documents. Specifically, the zipped data on \href{www.congressionalbills.org/download.html}{www.congressionalbills.org/download.html} that connects bills to labels contains numerical codes for the labels and URLs pointing to the bills.

On the other hand, since \texttt{Wiki} text-label mapping might have appeared in the pre-training corpus of LLMs, \frameworkname's performance on this dataset reflects a best-case scenario for our algorithm \cite{thompson_topic_2020}. However, even in this potential case, \frameworkname\ does not directly reproduce any ground truth topics from \texttt{Wiki} unless explicitly provided as example topics. Furthermore, the performance on \texttt{Wiki} is not perfect, indicating that simply memorizing ground truth topics cannot fully explain the results. Our novel prompting template, along with providing human-curated topic examples, means the model outputs are more guided by these few-shot demonstrations rather than resembling verbatim memorization. 
\newpage
\section{Additional seeded/anchored topic models}
\label{appendix:seeded-tm}
We include here the description and results of two additional seeded/anchored topics models: CorEx \cite{gallagher_anchored_2017} and BERTopic-guided \cite{grootendorst2022bertopic}. 

\paragraph{Seed words:} For all seeded topic models (CorEx, BERTopic-guided, and SeededLDA), we use the following seed words:
\begin{itemize}
    \item \texttt{Bills}: ('trade', 'practices', 'products') and ('agriculture', 'capital', 'goods', 'services').
    \item \texttt{Wiki}: ('natural', 'sciences', 'biology', 'chemistry', 'physics') and ('engineering', 'system', 'infrastructure').
\end{itemize}
\paragraph{CorEx:} CorEx \cite{gallagher_anchored_2017} dispenses with the generative assumptions of LDA and instead uses total correlation to learn topics. Anchored CorEx incorporates user-provided seed words to create topics that balance between the original corpus and the information from the seed topics.
\paragraph{BERTopic-guided:} BERtopic-guided is the seeded variant of BERTopic \cite{grootendorst2022bertopic}. BERTopic assigns user-provided seed words to some documents based on their embeddings' cosine similarity. These assignments are then combined with UMAP to create a self-supervised approach, nudging topic creation to the seed topics. 

Table \ref{appendix:seeded-tab} shows the topic alignment results for CorEx, SeededLDA, BERTopic-guided. Overall, CorEx and SeededLDA are slightly better than BERTopic-guided. Between CorEx and SeededLDA, CorEx is better when the number of topics is high ($k>100$ in \texttt{bills}). However, when $k<100$, SeededLDA outperforms CorEx. SeededLDA is comparable to LDA, but they all did not come close to TopicGPT. 

\begin{table*}[ht]
\small
\centering
\begin{tabular}{p{0.75cm}p{4.5cm}p{0.4cm}p{0.4cm}p{0.4cm}p{0.4cm}p{0.4cm}p{0.4cm}p{0.4cm}p{0.4cm}p{0.4cm}p{0.4cm}p{0.4cm}p{0.4cm}}
    \toprule
    \multirow{2}{*}{Dataset} & \multirow{2}{*}{Setting} & \multicolumn{3}{c}{TopicGPT} & \multicolumn{3}{c}{BERTopic-guided} & \multicolumn{3}{c}{CorEx} & \multicolumn{3}{c}{SeededLDA}\\
    \cmidrule(lr){3-5}\cmidrule(lr){6-8}\cmidrule(lr){9-11}\cmidrule(lr){12-14}
    & & $P_1$  & ARI & NMI  & $P_1$  & ARI & NMI & $P_1$ & ARI & NMI & $P_1$ & ARI & NMI\\
    \midrule
    \multirow{2}{*}{\texttt{Wiki}} & Default setting ($k$=31) & \textbf{0.73} & \textbf{0.58} & \textbf{0.71} & 0.26& 0.03& 0.21 & 0.52 & 0.3 & 0.46 & 0.61 & 0.47 & 0.65\\
    & Refined topics ($k$=22) & \textbf{0.74} & \textbf{0.60} & \textbf{0.70} & 0.26& 0.03& 0.19 & 0.51 & 0.27 & 0.46 & 0.62 & 0.51 & 0.65\\
    \midrule
    \multirow{2}{*}{\texttt{Bills}} & Default setting ($k$=79) & \textbf{0.57} & \textbf{0.42} & \textbf{0.52} & 0.30 & 0.05 & 0.35 & 0.38 & 0.19 & 0.31 & 0.50 & 0.28 & 0.43\\
     & Refined topics ($k$=24) & \textbf{0.57} & \textbf{0.40} & \textbf{0.49} & 0.26 & 0.04 & 0.22 & 0.37 & 0.17 & 0.29 & 0.52 & 0.31 & 0.45\\\midrule
     \multicolumn{14}{c}{\scriptsize{\emph{\frameworkname\ stability ablations, baselines controlled to have the same number of topics (\textbf{$k$}).}}}\vspace{0.1cm} \\
    \multirow{5}{*}{\texttt{Bills}} & Different generation sample ($k$=73) & \textbf{0.57} & \textbf{0.40} & \textbf{0.51} & 0.32& 0.07& 0.35 & 0.35& 0.16& 0.3 & 0.40 & 0.21 & 0.44\\
    & Out-of-domain prompts ($k$=147) & \textbf{0.55} & \textbf{0.39} & \textbf{0.51} & 0.27& 0.05& 0.37 & 0.38& 0.18& 0.33 & 0.29 & 0.13 & 0.44\\
    & Additional example topics ($k$=123) & \textbf{0.50} & \textbf{0.33} & \textbf{0.49} & 0.29& 0.05& 0.37 & 0.35& 0.17& 0.30 & 0.33 & 0.15 & 0.44 \\
    & Shuffled generation sample ($k$=118) & \textbf{0.55} & \textbf{0.40} & \textbf{0.52} & 0.30& 0.05& 0.37 & 0.42& 0.24& 0.34 & 0.34 & 0.18 & 0.44\\
    & Assigning with Mistral ($k$=79) & \textbf{0.51} & \textbf{0.37} & \textbf{0.47} & 0.30& 0.05& 0.35 & 0.38& 0.19& 0.31 & 0.50 & 0.28 & 0.43\\
    \bottomrule
\end{tabular}
\caption{\label{appendix:seeded-tab}Topical alignment between ground-truth labels and predicted assignments. Overall, \frameworkname\ achieves the best performance across all settings and metrics compared to SeededLDA, CorEx, and BERTopic-guided. The number of topics used in each setting is specified as $k$. The largest values in each metric and setting are \textbf{bolded}.}
\end{table*}
\newpage
\section{Prompts}
\label{appendix:prompts}
See Table \ref{tab:prompt-proposer-1}, \ref{tab:prompt-proposer-2}, \ref{tab:prompt-refiner}, \ref{tab:assigner} for the prompt templates that we used in our framework.
\begin{table*}[ht]
\small
\begin{tabular}{p{\linewidth}}
\toprule
\textbf{Prompt template for generating first-level/flat topics}
\\\midrule
You will receive a document and a set of top-level topics from a topic hierarchy. Your task is to identify generalizable topics within the document that can act as top-level topics in the hierarchy. If any relevant topics are missing from the provided set, please add them. Otherwise, output the existing top-level topics as identified in the document.\\
\vfill
[Top-level topics]\\
\textcolor{red}{\{Example topics (containing "\text{[1] Trade" in this example)}\}}\\
\vfill
[Examples]\\
Example 1: Adding "\text{[1]} Agriculture"\\
Document: \\
Saving Essential American Sailors Act or SEAS Act - Amends the Moving Ahead for Progress in the 21st Century Act (MAP-21) to repeal the Act's repeal of the agricultural export requirements that: (1) 25 of the gross tonnage of certain agricultural commodities or their products exported each fiscal year be transported on U.S. commercial vessels, and (2) the Secretary of Transportation (DOT) finance any increased ocean freight charges incurred in the transportation of such items. Revives and reinstates those repealed requirements to read as if they were never repealed.\\
\vfill
Your response: \\
\text{[1]} Agriculture: Mentions policies relating to agricultural practices and products.\\
\vfill

Example 2: Duplicate "\text{[1]} Trade", returning the existing topic\\
Document: \\
Amends the Harmonized Tariff Schedule of the United States to suspend temporarily the duty on mixtures containing Fluopyram.\\
\vfill
Your response: \\
\text{[1]} Trade: Mentions the exchange of capital, goods, and services.\\
\vfill
[Instructions]\\
Step 1: Determine topics mentioned in the document. \\
- The topic labels must be as GENERALIZABLE as possible. They must not be document-specific.\\
- The topics must reflect a SINGLE topic instead of a combination of topics.\\
- The new topics must have a level number, a short general label, and a topic description. \\
- The topics must be broad enough to accommodate future subtopics. \\
Step 2: Perform ONE of the following operations: \\
1. If there are already duplicates or relevant topics in the hierarchy, output those topics and stop here. \\
2. If the document contains no topic, return "None". \\
3. Otherwise, add your topic as a top-level topic. Stop here and output the added topic(s). DO NOT add any additional levels.\\

\vfill
[Document]\\
\textcolor{red}{\{Document\}}\\
\vfill
Please ONLY return the relevant or modified topics at the top level in the hierarchy.\\

[Your response]
\\\bottomrule
\end{tabular}
\caption{\label{tab:prompt-proposer-1}Prompt template for generating broad, high-level topics that can either serve as a flat list of standalone topics or as the first tier of a hierarchical topic taxonomy. Users should use this template in the typical, non-hierarchical use case. The designation of 'first-level' ensures these topics are sufficiently expansive to cover the topic distribution of the entire dataset. In practice, users need to modify the components highlighted in \textcolor{red}{red} (example topic list and document) as well as tailor the examples to their specific dataset.}
\end{table*}

\begin{table*}
\small
\begin{tabular}{p{\linewidth}}
\toprule
\textbf{Prompt template for generating second-level subtopics}
\\\midrule
You will receive a branch from a topic hierarchy along with some documents assigned to the top-level topic of that branch. Your task is to identify generalizable second-level topics that can act as subtopics to the top-level topic in the provided branch. Add your topic(s) if they are missing from the provided branch. Otherwise, return the existing relevant or duplicate topics. \\
\vfill
[Example] (Return "[2] Exports" (new) and "[2] Tariff" (existing) as the subtopics of "[1] Trade" (provided).)\\
Topic branch:\\
\text{[1]} Trade\\
\quad\text{[2]} Tariff\\
\quad\text{[2]} Foreign Investments\\
\vfill
Document 1: \\
Export Promotion Act of 2012 - Amends the Export Enhancement Act of 1988 to revise the duties of the Trade Promotion Coordinating Committee (TPCC). Requires the TPCC to: (1) make a recommendation for the annual unified federal trade promotion budget to the President; and (2) review the proposed fiscal year budget of each federal agency with responsibility for export promotion or export financing activities before it is submitted to the Office of Management and Budget (OMB) and the President, when (as required by current law) assessing the appropriate levels and allocation of resources among such agencies in support of such activities. \\
\vfill
Document 2: \\
Amends the Harmonized Tariff Schedule of the United States to suspend temporarily the duty on mixtures containing Fluopyram.\\
\vfill
Document 3: \\
Securing Exports Through Coordination and Technology Act - Amends the Foreign Relations Authorization Act, Fiscal Year 2003. Requires carriers obliged to file Shipper's Export Declarations to file them through AES (either directly or through intermediaries) before items are exported from any U.S. port, unless the Secretary of Commerce grants an exception. \\
\vfill
Your response: \\
\text{[1]} Trade\\
\quad\text{[2]} Exports (Document: 1, 3): Mentions export policies on goods.\\
\quad\text{[2]} Tariff (Document: 2): Mentions tax policies on imports or exports of goods. \\
\vfill
[Instructions]\\
Step 1: Determine PRIMARY and GENERALIZABLE topics mentioned in the documents. \\
- The topics must be generalizable among the provided documents. \\
- Each topic must not be too specific so that it can accommodate future subtopics.\\
- Each topic must reflect a SINGLE topic instead of a combination of topics.\\
- Each top-level topic must have a level number and a short label. Second-level topics should also include the original documents associated with these topics (separated by commas) as well as a short description of the topic.\\
- The number of topics proposed cannot exceed the number of documents provided.\\
Step 2: Perform ONE of the following operations: \\
1. If the provided top-level topic is specific enough, DO NOT add any subtopics. Return the provided top-level topic.\\
2. If your topic is duplicate or relevant to the provided topics, DO NOT add any subtopics. Return the existing relevant topic. \\
3. If your topic is relevant to and more specific than the provided top-level topic, add your topic as a second-level topic. DO NOT add to the first or third level of the hierarchy. \\
\vfill
[Topic branch]\\
\textcolor{red}{\{Topic\}}\\
\vfill
[Documents]\\
\textcolor{red}{\{Documents\}}\\
\vfill
DO NOT add first- or third-level topics.\\

[Your response]

\\\bottomrule
\end{tabular}
\caption{\label{tab:prompt-proposer-2}Prompt template for generating second-level subtopics in a topic hierarchy. In practice, users need to modify the components highlighted in \textcolor{red}{red} (generated topic branch and associated documents) as well as tailor the examples to their specific dataset. This prompt can be further accommodate subtopic generation for lower-level by changing the topic level in the examples and instructions.}
\end{table*}

\begin{table*}
\small
\begin{tabular}{p{\linewidth}}
\toprule
\textbf{Prompt template for refining (merging) topics}
\\\midrule
You will receive a list of topics that belong to the same level of a topic hierarchy. Your task is to merge topics that are paraphrases or near duplicates of one another. Return "None" if no modification is needed. \\
\vfill
\text{[Examples]} \\
Example 1: Merging topics ("[1] Employer Taxes" and "[1] Employment Tax Reporting" into "[1] Employment Taxes")\\
Topic List: \\
\text{[1]} Employer Taxes: Mentions taxation policy for employer\\
\text{[1]} Employment Tax Reporting: Mentions reporting requirements for employer\\
\text{[1]} Immigration: Mentions policies and laws on the immigration process\\
\text{[1]} Voting: Mentions rules and regulation for the voting process\\
\vfill
Your response: \\
\text{[1]} Employment Taxes: Mentions taxation report and requirement for employer (\text{[1]} Employer Taxes, \text{[1]} Employment Tax Reporting)\\

\vfill
Example 2: Merging topics (\text{[2]} Digital Literacy and \text{[2]} Telecommunications into \text{[2]} Technology)\\
\text{[2]} Mathematics: Discuss mathematical concepts, figures and breakthroughs. \\
\text{[2]} Digital Literacy: Discuss the ability to use technology to find, evaluate, create, and communicate information.\\
\text{[2]} Telecommunications: Mentions policies and regulations related to the telecommunications industry, including wireless service providers and consumer rights.\\
\vfill
Your response\\
\text{[2]} Technology: Discuss technology and its impact on society. (\text{[2]} Digital Literacy, \text{[2]} Telecommunications)\\
\vfill
[Rules]\\
- Each line represents a topic, with a level indicator and a topic label. \\
- Perform the following operations as many times as needed: \\
    - Merge relevant topics into a single topic.\\
    - Do nothing and return "None" if no modification is needed.\\
- When merging, the output format should contain a level indicator, the updated label and description, followed by the original topics.\\
\vfill
[Topic List]\\
\textcolor{red}{\{Topics\}}\\
\vfill
Output the modification or "None" where appropriate. Do not output anything else.\\

[Your response]
\\\bottomrule
\end{tabular}
\caption{\label{tab:prompt-refiner}Prompt template for refining/merging similar topics in a topic hierarchy. In practice, users need to modify the components highlighted in red (example topic list and document) to contain the similar topic pairs.}
\end{table*}

\begin{table*}
\small
\begin{tabular}{p{\linewidth}}
\toprule
\textbf{Prompt template for assigning topics}
\\\midrule
You will receive a document and a topic hierarchy. Assign the document to the most relevant topics the hierarchy. Then, output the topic labels, assignment reasoning and supporting quotes from the document. DO NOT make up new topics or quotes.  \\
\vfill
Here is the topic hierarchy:\\
\textcolor{red}{\{tree\}}\\
\vfill
[Examples]\\
Example 1: Assign "[1] Agriculture" to the document\\
Document: \\
Saving Essential American Sailors Act or SEAS Act - Amends the Moving Ahead for Progress in the 21st Century Act (MAP-21) to repeal the Act's repeal of the agricultural export requirements that: (1) 25\% of the gross tonnage of certain agricultural commodities or their products exported each fiscal year be transported on U.S. commercial vessels, and (2) the Secretary of Transportation (DOT) finance any increased ocean freight charges incurred in the transportation of such items.\\
\vfill
Your response:\\
\text{[1]} Agriculture: Mentions changes in agricultural export requirements ("...repeal of the agricultural export requirements that...")\\
\vfill
Example 2: Assign "[2] Tariff" to the document\\
Document: \\
Amends the Harmonized Tariff Schedule of the United States to suspend temporarily the duty on mixtures containing Fluopyram.\\
\vfill
Your response: \\
\text{[1]} Trade\\
\quad \text{[2]} Tariff: Mentions adjusting the taxation on mixtures containing Fluopyram ("...suspend temporarily the duty on mixtures containing Fluopyram.")\\
\vfill
[Instructions]\\
1. Topic labels must be present in the provided topic hierarchy. You MUST NOT make up new topics. \\
2. The quote must be taken from the document. You MUST NOT make up quotes. \\
3. If the assigned topic is not on the top level, you must also output the path from the top-level topic to the assigned topic.\\
\vfill
[Document] \\
\textcolor{red}{\{Document\}}\\
\vfill
Double check that your assignment exists in the hierarchy!\\

[Your response]

\\\bottomrule
\end{tabular}
\caption{\label{tab:assigner}Prompt template for assigning topics to a given document in the corpus. In practice, users need to modify the components highlighted in \textcolor{red}{red} as well as tailor the examples to the specific dataset (can be reused from prompts in previous stages). Users can also modify the prompt to strictly assign to one topic.}
\end{table*}

\begin{table*}
\centering
\small
\begin{tabularx}{\textwidth}{p{.05\textwidth}Xp{.1\textwidth}p{.1\textwidth}p{.1\textwidth} }

 \toprule
 Dataset & Document & Ground-truth label & Initial label & Reassigned labels \\
 \midrule
 \texttt{Bills} & Securing Health for Ocean Resources and Environment Act or the SHORE Act - Requires the Under Secretary for Oceans and Atmosphere to: (1) review the National Oceanic and Atmospheric Administration's (NOAA) capacity to respond to oil spills; (2) be responsible for developing and maintaining oil spill trajectory modeling capabilities... & Technology & Environment & Technology; Environment \\
 \midrule
 \texttt{Bills} & Driver Fatigue Prevention Act. This bill amends the Fair Labor Standards Act of 1938 to apply its maximum hours requirements to over-the-road bus drivers. & Labor & Transportation Safety & Labor; Transportation Safety\\
 \midrule
 \texttt{Bills} & Defense Travel Simplification Act of 2007 - Requires the Secretary of Defense to: (1) redesignate the Defense Travel System as the Defense Travel Accounting and Voucher Processing System; and (2) establish an intra-agency task force to recommend measures to streamline and simplify the commercial travel system... & Domestic Commerce & State and Local Government & Technology; Military and Veterans Affairs\\
 \midrule
 \texttt{Bills} & Amends the Internal Revenue Code to allow until June 30, 2010: (1) a first-time homebuyer tax credit for all purchasers of a principal residence (not just first-time homebuyers); and (2) a refundable tax credit, up to \$3,000, for the costs of refinancing a principal residence. & Domestic Commerce & Housing & Housing \\
 \midrule
 \texttt{Bills}& Amends the Internal Revenue Code to extend through 2014 the equalization of the exclusion from gross income for employer-provided mass transit and parking benefits. & Labor & Transportation Safety & Taxation; Transportation Safety\\
 \midrule
 \texttt{Wiki} & Colonel Cyrus Kurtz Holliday (April 3, 1826 – March 29, 1900) was one of the founders of the township of Topeka, Kansas, in the mid 19th century; and was Adjutant General of Kansas during the American Civil War. The title Colonel, however, was honorary. He was the first president of the Atchison, Topeka and Santa Fe Railway, as well as one of the railroad's directors for nearly 40 years, up to 1900... & Engineering and Technology & History and Politics & History and Politics; Engineering and Technology\\
  \midrule
  \texttt{Wiki} & Jack Banham Coggins (July 10, 1911 – January 30, 2006) was an artist, author, and illustrator. He is known in the United States for his oil paintings, which focused predominantly on marine subjects. He is also known for his books on space travel, which were both authored and illustrated by Coggins. Besides his own works, Coggins also provided illustrations for advertisements, magazine covers and articles.... & Language and literature & Art and Craftmanship & Art and Craftmanship; Literature and Writing\\
  \midrule
  \texttt{Wiki} & HMS Belfast is a museum ship, originally a Royal Navy light cruiser, permanently moored in London on the River Thames and operated by the Imperial War Museum. Construction of Belfast, the first Royal Navy ship to be named after the capital city of Northern Ireland, and one of ten Town-class cruisers, began in December 1936... &  Warfare & History and Politics & Military and Warfare \\
  \midrule
  \makecell[l]{\texttt{Wiki}} & The Grand Street Bridge was a double-leaf deck-girder bascule bridge in Bridgeport, Connecticut, United States, that spanned the Pequonnock River and connected Grand Street and Artic Street. It was one of three movable bridges planned by the City of Bridgeport in 1916 at the request of the War Department during World War I..... & Art and architecture & Engineering and Technology & Engineering and Technology\\
  \midrule
  \makecell[l]{\texttt{Wiki}} & Burger King Specialty Sandwiches = The Burger King Specialty Sandwiches are a line of sandwiches developed by the international fast-food restaurant chain Burger King in 1978 and introduced in 1979 as part of a new product line designed to expand Burger King's menu with more sophisticated, adult-oriented fare beyond hamburgers... & Agriculture, food, and drink & Business and Finance & Food and Cooking; Business and Finance; Advertising and Marketing\\
 \bottomrule
\end{tabularx}
\caption{\label{tab:result-error-analysis} Error analysis on five examples from each of \texttt{Bills} and \texttt{Wiki} datasets. Documents are truncated for ease of viewing.}
\end{table*}

%% file: sections/5b-Expert.tex
As an additional form of evaluation, we conducted a case study directly comparing \frameworkname\ with categories in a different domain \citep{resnik:rtl}. These categories were obtained using TOPCAT, a qualitative content analysis protocol involving structured human curation of automatically obtained topical categories \citep{topcat:tada,resnik:topcat}.\footnote{Selecting the model size (number of topics) is part of the protocol.} The dataset comprised more than 16K responses to a question on Reddit asking formerly suicidal Redditors what had gotten them through their dark times.\footnote{\url{https://www.reddit.com/r/AskReddit/comments/j0z4lp/formerly_suicidal_redditors_whats_something_that/}. This work received an Exempt determination from our Institutional Review Board.} A 30-topic LDA model, created using preprocessing essentially identical to the preprocessing described in Section~\ref{sec:baselines}, served as input to curation by experts in the study of and/or prevention of suicide, resulting in the 24~categories in Table~\ref{tbl:rtlexpert}.\footnote{\citet{resnik:rtl} situates these categories in relationship to theories of suicide, contrasts them with Linehan's (\citeyear{linehan1983reasons}) widely used Reasons for Living inventory, and discusses potential clinical implications.} \frameworkname\ produced the categories in Table~\ref{tbl:rtl\frameworkname}.

Qualitatively, \frameworkname\ has succeeded in capturing many aspects of the expert-derived category analysis. Denoting expert categories as E\emph{i} and \frameworkname\ categories as T\emph{j}, we see:
\begin{itemize}
\item Direct correspondences. These include T1 with E8, T6 with E21, T11 with E4, and arguably T8 with E10 and T14 with E7.
\item Cross-cutting but reasonable categories.  Categories T4, T12, and T13 all  bear a clearly visible relationship to E1--E5. All of these represent variations on a general theme of concern about others who are left behind, which is the essence of the experts' first metacategory, although the experts and \frameworkname\ carve up the space in different ways.
\item Higher level themes. This includes T5, which (ignoring the oddly specific example in the description) is a higher-level category in relation to E16, E17, E18, E20, and arguably E19. It may also be an alternative characterization of cross-cutting categories.
\end{itemize}
There are also some categories that were not among the expert-derived categories. These all appear to capture ideas that are valid in terms of characterizing themes present for many responses in the document collection but are not directly responsive to the question: T2, T3, and T10 all appear to characterize background content rather than categories of responses \emph{per se} to the question that was asked, and T9 characterizes non-answers. This highlights the fact that qualitative content analysis is never driven just by the data; it is also driven by the research questions behind the data analysis \citep{topcat:tada,resnik:topcat,resnik:rtl}. We would argue that these \frameworkname\ categories can nonetheless be useful, both because they potentially provide insight into themes about background information that could potentially be correlated with reasons individuals chose not to die by suicide, and because having well-defined "irrelevant" categories, particularly T9 ("junk"responses), may contribute to useful ways to filter and/or further break down the collection of responses.

\begin{table*}
\small
\begin{tabular}{p{\linewidth}}
\toprule
\textbf{Expert "reasons to live" categories}
\begin{enumerate}[label=E\arabic*]
\item Not wanting to hurt specific family members
\item Financial burdens for survivors or if attempt fails
\item Not wanting to traumatize loved ones, especially finding and having to deal with the body
\item Pets, especially cats and dogs -- not wanting to hurt them, worry about them, etc.
\item Not wanting to hurt specific family members, particularly children
\item[]
\item Holding on for one more day and focusing on the next thing, routine, sleep on it and see if/how things get better in the morning
\item Quotes, inspiration, philosophy, wonder -- things that resonate/connect
\item Sense of connection -- feeling seen, heard, understood, cared about by other people 
\item Professional diagnosis and treatment
\item Hope and connection with and/or fear, from spiritual or religious beliefs
\item Recognition that suicide is "a permanent solution to a temporary problem" and life will change
\item Hope -- looking past the bad and feeling like life can get better
\item Insight/realization about not wanting to die or finding meaning in surviving an attempt
\item Professional diagnosis and treatment 
\item Insights from reading or writing, including writing a suicide note
\item[] 
\item Looking forward to next episodes of TV, new games, experiencing new media
\item Music -- wanting to continue hearing it, experiencing music as meaningful, specific artists
\item Food and sensory pleasures
\item Substance use and abuse
\item Spending time on alternative media
\item Turning point, transition, new connection
\item Finding a new distraction or focus
\item[]
\item Fear of surviving an attempt and being worse off
\item Spite -- proving people wrong
\end{enumerate}
\\\bottomrule
\end{tabular}
\caption{\label{tbl:rtlexpert}Categories obtained using a topic-oriented protocol  for content analysis of text, applied to more than 16,000 responses from formerly suicidal individuals about what got them through their dark times \citep{resnik:rtl}. Two subject matter experts independently followed a structured series of steps to curate a 30-topic LDA model, followed by a consensus process yielding 24~relevant categories. Further expert analysis grouped these into four themes or meta-categories.}
\end{table*}

\begin{table*}
\small
\begin{tabular}{p{\linewidth}}
\toprule
\textbf{\frameworkname\ "reasons to live" categories}
\begin{enumerate}[label=T\arabic*]
\item Emotionally supportive close relationships: The respondent cites emotional support from family, a significant other, or friend.
\item Emotional distress: The respondent expresses feelings of emotional distress, including feelings of being unappreciated, belittled, and harassed.
\item Social alienation: The respondent feels alienated or ostracized due to differing opinions or beliefs.
\item Fear of causing harm to others: The respondent is concerned about the negative impact their death would have on their loved ones.
\item Therapeutic hobbies: The respondent finds solace and joy in sharing pictures of corgis, which helps them cope with their emotional distress.
\item Personal growth and goal setting: The respondent finds purpose and direction in setting personal goals and working towards them, such as pursuing a career in veterinary medicine.
\item Seeking professional help: The respondent mentions seeking help from a professional, such as a therapist or counselor.
\item Fear of the unknown: The respondent expresses uncertainty or fear about what happens after death.
\item Junk response: The respondent does not reply to the prompt or does not take the response seriously.
\item Loss of a loved one: The respondent is dealing with the death of someone close to them.
\item Responsibility towards pets: The respondent mentions the need to care for a pet, which includes providing food and affection.
\item Death would be a burden: The respondent believes that their death would have caused distress to others.
\item Responsibility towards others: The respondent feels obligated to keep a promise made to someone else.
\item Appreciation of life: The respondent expresses a deep appreciation for life and its beauty.
\end{enumerate}
\\\bottomrule
\end{tabular}
\caption{\label{tbl:rtl\frameworkname}Categories obtained automatically using \frameworkname\ for the same responses from formerly suicidal individuals about what got them through their dark times, for comparison with the expert categories in Table~\ref{tbl:rtlexpert}.}
\end{table*}

\begin{table*}
\centering
\small
    \begin{tabular}{llcccccccccccc}
        \toprule
         &  \frameworkname& SeededLDA\\
         \midrule
         Out-of-scope&  25\%& 25\%\\
         Missing&  \textbf{47.82}\%& 56.52\%\\
         Repeated&  21.43\%& 21.43\%\\
         \bottomrule
     \end{tabular}
    \caption{\label{tab:expert_eval} Comparison of misaligned topic proportions between SeededLDA ($k=14$) and unrefined \frameworkname\ outputs on Reason to Live dataset. Values are averaged over three annotations and rounded to two decimal places. The best (lower) misalignment proportion is \textbf{bolded}.}
\end{table*}

We further implement our human evaluation protocol (Section \ref{sec:human-eval}) on the dataset with two subject matter experts recruited via Upwork. Table \ref{tab:expert_eval} shows the average percentage of missing, out-of-scope, and repeated topics generated by \frameworkname\ and SeededLDA relative to the expert-curated ground truth. Unlike the protocol in Section \ref{sec:human-eval}, we compare with SeededLDA instead of LDA because LDA is used to craft the ground-truth expert categories, which may introduce bias. While \frameworkname\ and SeededLDA have similar proportions of out-of-scope and repeated topics, \frameworkname\ has fewer missing topics, indicating broader coverage. Both experts note that \frameworkname\ is superior to SeededLDA in terms of usability and interpretability, particularly because \frameworkname\ outputs follow a format that is coherent and ``more closely related to standardized codes in thematic analysis".